\documentclass{article}

\usepackage{microtype}
\usepackage{graphicx}
\usepackage{subcaption}
\usepackage{booktabs} % for professional tables
\usepackage{multirow}

\usepackage{hyperref}
\usepackage[most]{tcolorbox}

\usepackage[accepted]{icml2026}
\usepackage{xurl}
\usepackage{amsmath}
\usepackage{amssymb}
\usepackage{mathtools}
\usepackage{amsthm}
\usepackage{listings}
\definecolor{formbg}{RGB}{240,247,255}

\definecolor{keywordcolor}{rgb}{0.8, 0.0, 0.4} 

\usepackage[capitalize,noabbrev]{cleveref}

\theoremstyle{plain}

\theoremstyle{definition}

\theoremstyle{remark}

\usepackage[disable,textsize=tiny]{todonotes}
\icmltitlerunning{Extracting Interpretable Algorithms from the Discrete Transformer}

\begin{document}

\twocolumn[
  \icmltitle{Weights to Code: Extracting Interpretable Algorithms from \\
  the Discrete Transformer}



  \icmlsetsymbol{intern}{${\dagger}$}

  \begin{icmlauthorlist}
    \icmlauthor{Yifan Zhang}{lab,sch,intern}
    \icmlauthor{Wei Bi}{comp}
    \icmlauthor{Kechi Zhang}{lab,sch}
    \icmlauthor{Dongming Jin}{lab,sch}
    \icmlauthor{Jie Fu}{comp1,sch1}
    \icmlauthor{Zhi Jin}{lab,sch}
  \end{icmlauthorlist}

  % \icmlaffiliation{lab}{Key Laboratory of High Confidence Software Technology (PKU), MOE, China}
  % \icmlaffiliation{sch}{School of Computer Science, Peking University, China}
  % \icmlaffiliation{comp}{Kuaishou Technology}
  % \icmlaffiliation{comp1}{Shanghai AI Lab}
  % \icmlaffiliation{sch1}{Shanghai Innovation Institute}
  % 建议将具体的城市和国家补充完整
  \icmlaffiliation{lab}{Key Laboratory of High Confidence Software Technology (PKU), MOE, Beijing, China}
  \icmlaffiliation{sch}{School of Computer Science, Peking University, Beijing, China}
  \icmlaffiliation{comp}{Kuaishou Technology, Beijing, China}
  \icmlaffiliation{comp1}{Shanghai AI Lab, Shanghai, China}
  \icmlaffiliation{sch1}{Shanghai Innovation Institute, Shanghai, China}

  % \icmlcorrespondingauthor{Yifan Zhang}{yifanzhang@stu.pku.edu.cn}
  \icmlcorrespondingauthor{Jie Fu}{fujie@pjlab.org.cn}
  \icmlcorrespondingauthor{Zhi Jin}{zhijin@pku.edu.cn}

  \icmlkeywords{Machine Learning, ICML}

  \vskip 0.3in
]



% \printAffiliationsAndNotice{Work done during an internship at Kuaishou Technology.}
\printAffiliationsAndNotice{$^{\dagger}$\,Work done during an internship at Kuaishou Technology.}

\begin{abstract}
    Algorithm extraction aims to synthesize executable programs directly from models trained on algorithmic tasks, enabling \textit{de novo} recovery of executable mechanisms from weights without relying on human-written target programs.
    However, applying this paradigm to Transformer is complicated by representation entanglement (e.g., superposition), where features encoded in overlapping directions substantially hinder the recovery of symbolic expressions.
    We propose the Discrete Transformer, an architecture explicitly designed to bridge the gap between continuous representations and discrete symbolic logic.
    By injecting discreteness through temperature-annealed sampling, our framework effectively leverages hypothesis testing and symbolic regression to extract human-readable programs.
    Empirically, the Discrete Transformer achieves performance comparable to the RNN-based MIPS baseline on shared discrete tasks, while broadening extraction to tasks with continuous-valued intermediate computations.
    Finally, we show that architectural inductive biases provide fine-grained control over synthesized programs, establishing the Discrete Transformer as a controllable testbed for algorithm extraction and Transformer interpretability.
\end{abstract}

\section{Introduction} \label{Sec: Intro}

\begin{figure*}[ht] 
    \centering
    \includegraphics[width=\linewidth]{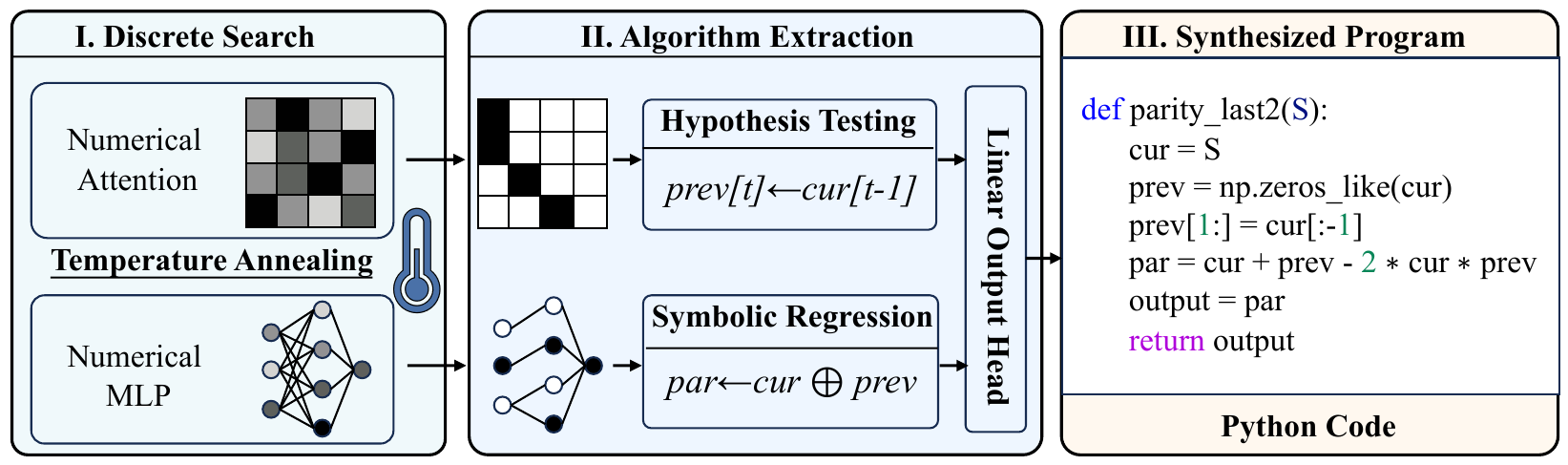}
    \caption{
    Illustration of the proposed framework for extracting executable algorithms from a Discrete Transformer. (I) Discrete Search: Temperature annealing is leveraged to encourage interpretable discretization in both Numerical Attention and MLP modules. (II) Algorithm Extraction: Attention patterns are characterized via hypothesis testing, while arithmetic transformations are approximated through symbolic regression. (III) Synthesized Program: The extracted components are integrated via a linear output head to generate Python code. As shown, the framework successfully recovers the \texttt{parity\_last2} algorithm, correctly implementing the arithmetic XOR logic.}
    \label{fig:overview}
\end{figure*}

Program synthesis is the task to construct a program that provably satisfies a given high-level formal specification, a long-standing problem which can date back to \citet{church1960application}.
In recent years, this field has been revolutionized by Large Language Models (LLMs), which have achieved remarkable success in code generation~\cite{rozière2024codellamaopenfoundation, guo2024deepseekcoderlargelanguagemodel, team2024codegemma}. 
Despite their success, an alternative paradigm---\emph{algorithm extraction}---offers a distinct advantage: the ability to recover executable mechanisms \emph{de novo} from trained model weights, without requiring human-written target programs. 
Recent work has demonstrated the feasibility of this approach in Recurrent Neural Networks (RNNs) via the Mechanistic-Interpretability-based Program Synthesis (MIPS), which successfully leverages symbolic regression to synthesize programs that accurately match neural network behavior~\cite{michaud2024openingaiblackbox}.

However, extending algorithm extraction to the dominant Transformer architecture faces significant challenges, rooted in a fundamental gap between the continuous, high-dimensional nature of Transformer and the discrete, sparse nature of symbolic algorithms.
The primary obstacle lies in interpreting the Transformer's internal representations.
Specifically, the standard Transformer often exhibits ``superposition'', where features are encoded in an overlapping, non-orthogonal set of directions rather than individual neurons~\cite{cunningham2023sparseautoencodershighlyinterpretable, elhage2022toy}.
Consequently, the model’s internal representations can be entangled and polysemantic. Unlike the disentangled input-output mappings required for symbolic regression, these representations make the direct extraction of explicit symbolic expressions challenging.
This motivates our central question: \textit{Is it possible to synthesize executable and interpretable programs by extracting algorithms from Transformer?}
% Although \citet{friedman2023learning} have introduced Transformer Programs to translate weights into code, we argue that such translations primarily reproduce forward computation, i.e. \emph{simulating} execution rather than \emph{abstracting} a general, human-readable algorithm.

% Extending algorithm extraction from RNNs to Transformers faces significant challenges, rooted in a fundamental gap between the continuous, high-dimensional nature of Transformer and the discrete, sparse nature of symbolic algorithms.
% The primary obstacle lies in interpreting the Transformer's internal representations.
% Specifically, the standard Transformer often exhibits ``superposition'', where features are encoded in an overlapping, non-orthogonal set of directions rather than individual neurons \cite{cunningham2023sparseautoencodershighlyinterpretable, elhage2022toy}.
% Consequently, the model’s internal representations are highly entangled and polysemantic. Unlike the disentangled input-output mappings required for symbolic regression, these representations render the direct application extraction of explicit symbolic expressions infeasible.

In this work, we propose the \emph{Discrete Transformer}, an architecture explicitly optimized for algorithm extraction (see Figure~\ref{fig:overview}).
Building upon the framework of Transformer Programs~\cite{friedman2023learning}, our design incorporates critical modifications to facilitate the transition from continuous dynamics to discrete logic.
Architecturally, the Discrete Transformer comprises the Numerical Attention, Numerical MLP, and linear output head.
Aligning with the Restricted Access Sequence Processing (RASP) computational model~\cite{weiss2021thinking}, we impose a strict functional disentanglement, where the Numerical Attention is responsible for routing information across positions, while the Numerical MLP is dedicated solely to element-wise arithmetic operations.
Crucially, we incorporate differentiable sampling mechanisms into each module to inject temperature-controlled discreteness. 
Through annealing, the model gradually transitions into a fully discrete computation graph, ensuring that the solution to the algorithmic task is implicitly but clearly encoded within its weights.

Once the model converges to a discrete state, we employ a modular strategy to recover the underlying algorithm. 
Recognizing the distinct roles of the components, we adopt \textit{hypothesis testing} for the Numerical Attention modules to identify interpretable routing patterns, and apply \textit{symbolic regression} to the Numerical MLP modules to infer the specific arithmetic expressions. 
Finally, these extracted primitives are aggregated through the linear output head, yielding a concise, human-readable, and executable program that verifiably solves the target task.

We validate our approach across a diverse suite of algorithmic tasks, achieving comparable algorithm extraction performance to the RNN-based prior method (MIPS).
Crucially, unlike MIPS, which is constrained by the finite-state approximations imposed by post-hoc integer autoencoders, our Discrete Transformer natively processes continuous latent variables, thereby substantially broadening the scope of mechanistic-interpretability-based program synthesis.
% Crucially, unlike the MIPS restricted by discrete latent variable approximations, our Discrete Transformer natively processes continuous latent variables, substantially broadening the scope of mechanistic-interpretability-based program synthesis. 
Beyond performance, we provide empirical analyses of the framework's architectural sensitivity, training dynamics, and controllability.
In addition to revealing that functional convergence precedes full structural discretization~\cite{louizos2017learning, savarese2021winninglotterycontinuoussparsification}, we demonstrate that tailoring architectural constraints imposes strong inductive biases, establishing the Discrete Transformer as a controllable framework for interpretable algorithm discovery.
% In summary, our findings demonstrate that the Discrete Transformer offers a principled framework for converting the opaque statistical patterns of Transformer into executable, human-interpretable algorithms.
% Bridging program synthesis with differentiable structure learning \cite{louizos2017learning, savarese2021winninglotterycontinuoussparsification}, we reveal that functional convergence typically precedes full structural discretization.
% Moreover, by leveraging architectural constraints to limit the solution space, we validate the Discrete Transformer as a controllable framework for algorithm discovery.
% Conceptually bridging program synthesis with differentiable structure learning and continuous sparsification \cite{louizos2017learning, savarese2021winninglotterycontinuoussparsification}, we reveal that the model typically achieves functional convergence, indicated by a rapid decrease in loss, prior to the full structural discretization enforced by temperature annealing. 
% Furthermore, we demonstrate that by explicitly manipulating architectural constraints and training configurations, we can impose strong inductive biases to constrain the solution space, establishing the Discrete Transformer as a controllable framework for interpretable algorithm discovery.

Overall, our main contributions are as follows:
\begin{itemize} 
\item We propose the Discrete Transformer, a RASP-aligned architecture that enforces functional disentanglement to bridge continuous optimization and discrete logic.
\item We develop a modular extraction pipeline combining hypothesis testing and symbolic regression, extending mechanistic-interpretability-based synthesis to continuous variables.
\item We provide empirical analyses of capacity sensitivity, training dynamics, and architectural interventions, showing how inductive biases can steer the extraction toward alternative but equivalent executable programs.
\end{itemize}

\paragraph{Conflict of Interest Disclosure.}
Y.Z.\ proposed and developed the Discrete Transformer during an internship at Kuaishou Technology, and W.B.\ is employed by Kuaishou Technology. The Discrete Transformer is a research model proposed and evaluated in this paper. The authors declare no other financial or substantive conflicts of interest related to this work.

\section{Related Work} \label{Sec: Related Work}

\subsection{Mechanistic Interpretability in Transformer}

The foundation for connecting Transformer with programs is the RASP \cite{weiss2021thinking}, which abstracts sequence processing into primitives.
% where \texttt{select-aggregate} operations correspond to attention mechanisms and element-wise operations correspond to MLPs. 
Building on this abstraction, Tracr~\cite{lindner2023tracr} compiles such programs \textit{into} Transformer weights, whereas Transformer Programs~\cite{friedman2023learning} and Adaptive Transformer Programs~\cite{ICLR2025_9d560961} tackle the inverse problem of \textit{learning} interpretable structures. 
By imposing strict architectural constraints and employing discrete optimization techniques, such as Gumbel-Softmax~\cite{jang2016categorical}, to promote discretization, these approaches produce models that can be deterministically mapped to programs.

While closely related to our work, Transformer Programs and Adaptive Transformer Programs differ significantly in both representation and objective. 
Specifically, these frameworks translate forward computations into bounded symbolic formalisms, typically constrained to finite grids or categorical variables. 
In contrast, our approach recovers compact, executable algebraic programs directly from trained weights via symbolic regression. 
Furthermore, their focus is faithful forward decompilation, while ours is abstract program recovery from learned mechanisms. 
Consequently, we regard them as related yet conceptually distinct baselines.

% Despite the resulting executable programs, their representations are inherently bounded (e.g., finite-grid or categorical logic) and thus do not naturally capture open-form algebraic logic targeted by our symbolic regression–based extraction.
% However, we argue that such direct translation merely simulates low-level operations on test cases; true algorithm extraction requires explicit symbolic reasoning to distill abstract, high-level algorithm logic from these discrete components.

\subsection{Symbolic Regression and Algorithm Extraction}

Symbolic regression searches for closed-form expressions balancing accuracy and complexity~\cite{udrescu2020aifeynman20paretooptimal, cranmer2023interpretablemachinelearningscience}. While traditionally used for scientific discovery~\cite{cranmer2020discoveringsymbolicmodelsdeep, ma2022evolving} or recently enhanced by LLMs' scientific priors~\cite{shojaee2024llm}, we leverage it as the core engine for algorithm extraction—synthesizing concise, readable programs directly from trained neural networks.

The innovative MIPS framework \cite{michaud2024openingaiblackbox} pioneers this approach for RNNs. Analogous to the use of Sparse Autoencoders (SAEs) in interpreting features \cite{cunningham2023sparseautoencodershighlyinterpretable, bricken2023towards}, the MIPS employs post-hoc integer autoencoders to approximate continuous latent states into finite states suitable for symbolic regression. 
% Our work adopts this ``extraction via symbolic regression'' paradigm but eliminates the need for such post-hoc interpretation. 
% Unlike MIPS, which depends on separate quantization modules to bridge the gap between continuous and discrete representations, 
In contrast to the MIPS, which relies on auxiliary quantization modules to approximate discreteness, our Discrete Transformer is architecturally designed to learn an interpretable, inherently discrete computation graph, thereby enabling direct and seamless algorithm extraction.

\section{Discrete Transformer} \label{Sec: Discrete Transformer}

In this section, we introduce the Discrete Transformer, a specialized architecture purposefully designed for symbolic regression and algorithm extraction. 
% Deviating from the standard Transformer, where information is often entangled within high-dimensional latent vectors~\cite{vaswani2023attentionneed}, our variant operates on a disentangled residual stream of scalar variables. 
It is structured as a computational graph with clear functional specialization: the Numerical Attention performs explicit variable routing, while the Numerical MLP is responsible for arithmetic computation.
% Specifically, each module directs its scalar output into a dedicated, orthogonal subspace. Furthermore, the architecture is engineered to enforce interpretability by facilitating a smooth, differentiable transition from continuous representations to discrete symbolic logic.
% This design enforces a strict separation of concerns: the model is structurally constrained to function as a coherent computational graph, where dedicated modules perform explicit routing (Numerical Attention) or arithmetic (Numerical MLP) operations on distinct variables.
% This design enforces a strict functional specialization: by structurally constraining the model to operate as a coherent computational graph, we assign distinct roles to its components—Numerical Attention is dedicated to variable routing, while Numerical MLP focuses on arithmetic operations.
% Specifically, the architecture is structured as a computational graph, where Numerical Attention is dedicated to variable routing, while Numerical MLP focuses on arithmetic operations.

% \subsection{Discrete Transformer Framework}
% \label{subsec:discrete_transformer_framework}

\subsection{Numerical Residual Stream}
\label{subsec:numerical_residual_stream}

% The Discrete Transformer deviates from the standard Transformer architecture in two fundamental ways: the structure of the residual stream and the mechanism of information processing.
The essential difference between the Discrete Transformer and the standard one lies in the structure of the residual stream and the mechanism of information processing.
To align with the symbolic nature of algorithm extraction, we define the residual stream not as latent vectors, but as a collection of explicit scalar variables. 
Let $\mathbf{h}_l = [x_1, \dots, x_{N_l}]^\top \in \mathbb{R}^{N_l}$ denote the residual stream at layer $l$, consisting of $N_l$ distinct scalar variables.\footnote{Unless otherwise stated, equations in this section and the subsequent extraction procedure are applied position-wise and in parallel over sequence positions; only Numerical Attention operates across positions.}
In contrast to standard additive updates, we adapt the concatenation principle~\cite{friedman2023learning, ICLR2025_9d560961} to the scalar domain, updating the residual stream with new outputs $\mathbf{o}_l$:
\begin{equation}
\mathbf{h}_{l+1} = \text{Concat}(\mathbf{h}_l, \mathbf{o}_l) \in \mathbb{R}^{N_l + |\mathbf{o}_l|}.
\end{equation}
% In contrast to the standard additive residual update employed in the vector-valued Transformer ($\mathbf{h}_{l} = \mathbf{h}_{l-1} + f(\mathbf{h}_{l-1})$), which mixes newly computed features with existing ones in the same coordinates, we adopt a concatenation-based update rule inspired by \citealp{friedman2023learning, ICLR2025_9d560961}, yet rigorously constrained to scalar granularity. 
% Specifically, if layer $l$ generates a set of new scalar outputs $\mathbf{o}_{l} \in \mathbb{R}^{|\mathbf{o}_{l}|}$, the stream is updated as:
% \begin{equation}
%     \mathbf{h}_{l+1} = \text{Concat}(\mathbf{h}_l, \mathbf{o}_l) \in \mathbb{R}^{N_l + |\mathbf{o}_l|}.
% \end{equation}
% This design preserves all intermediate variables as separate coordinates, ensuring that the full computational history remains accessible and disentangled, and thereby alleviating the information interference typically induced by superposition in dense vector representations.
This design preserves the full computational history as disentangled coordinates, alleviating the information interference induced by superposition in dense vectors.

% To identify operands, we design a discretized reading mechanism inspired by learnable input selection~\cite{friedman2023learning}. We learn a projection $\mathbf{W}_{\text{read}} \in \mathbb{R}^{k \times N_{l}}$ and apply the smooth transition mechanism~\cite{ICLR2025_9d560961}—bridging Gumbel-Softmax~\cite{jang2016categorical} and Sparsemax~\cite{martins2016softmax}—to anneal logits. Inputs are retrieved via:
% \begin{equation}
% \mathbf{u} = S(\mathbf{W}_{\text{read}}, \tau) \cdot \mathbf{h}_l.
% \end{equation}
% As temperature $\tau \to 0$, the selection distribution converges to a deterministic one-hot indicator, effectively functioning as a differentiable pointer over the computation graph.

% \textbf{Discretized Input Selection.} 
Moreover, a critical challenge in algorithm extraction is to identify \textit{which} variables serve as operands. 
Inspired by learnable input selection~\cite{friedman2023learning}, we address this by designing a discretized reading mechanism tailored for the numerical residual stream: for any computational module (Numerical Attention or MLP) at layer $l$ requiring $k$ inputs, we learn a projection matrix $\mathbf{W}_{\text{read}} \in \mathbb{R}^{k \times N_{l}}$ to select inputs from the residual stream $\mathbf{h}_l$.
% , which is defined by $k$ indictors $\pi_k \in \{0, 1\}^{N_{in}}$.
% To enable discrete structure optimization within a differentiable framework, we adopt the smooth transition mechanism~\cite{ICLR2025_9d560961}, which bridges the Gumbel-Softmax~\cite{jang2016categorical} and the Sparsemax~\cite{martins2016softmax} to anneal the projection matrix.
To enable discrete structure optimization within a differentiable framework, we apply a temperature-controlled sampling function $S(\cdot, \tau)$ (derivation provided in Appendix~\ref{sec: smooth transition mechanism}) row-wise to the logits $\mathbf{W}_{\text{read}}$.
The input vector $\mathbf{u} \in \mathbb{R}^k$ is then retrieved from the current residual stream $\mathbf{h}_l \in \mathbb{R}^{N_l}$ via:
\begin{equation}
\mathbf{u} = S(\mathbf{W}_{\text{read}}, \tau) \mathbf{h}_l,
\end{equation}
where $\tau$ is the annealing temperature.
As temperature $\tau \to 0$, the selection distribution converges to a deterministic one-hot indicator, effectively functioning as a differentiable pointer over the computation graph.
% Specifically, we apply a temperature-controlled sampling function $S(\cdot, \tau)$ row-wise to $\mathbf{W}_{\text{read}}$. 
% The input vector $\mathbf{u} \in \mathbb{R}^k$ for the module is obtained via:
% \begin{equation}
% \mathbf{u} = S(\mathbf{W}_{\text{read}}, \tau) \cdot \mathbf{h_l},
% \end{equation}
% where $\mathbf{h}_l \in \mathbb{R}^{N_l}$ is the current residual stream and $\tau$ is the annealing temperature.
% As $\tau \to 0$, $S(\cdot)$ approaches a hard one-hot selection, effectively pointing to specific indices in $\mathbf{h}_l$. 
% As the temperature $\tau \to 0$, the continuous relaxation converges to deterministic one-hot indicators $\pi_k \in \{0, 1\}^{N_{l}}$.
% This mechanism acts as a differentiable pointer, allowing the model to learn the topology of the computation graph.
% This mechanism functions as a differentiable pointer, enabling the model to uncover the underlying topology of the computation graph.

\subsection{Numerical Attention as Router}
\label{subsec:numerical_attention}

Following the architectural paradigm established in Tracr~\cite{lindner2023tracr} and Transformer Programs~\cite{friedman2023learning}, we employ a hard attention mechanism designed strictly as an information routing operator, rather than a feature mixer.
% We adopt a constraint from Tracr \cite{lindner2023tracr} and Transformer Programs \cite{friedman2023learning}, and design a hard attention mechanism to serve strictly as an information routing operator rather than a feature mixer. 
In the context of algorithm extraction, this design can help isolate all computational transformations within the Numerical MLP, thereby reducing the difficulty of attention analysis.
% Since conducting symbolic regression on complex attention is challenging, we adopt this design to consolidate data transformations within simpler MLP, thereby reserving the extraction of expressions via symbolic regression exclusively for the MLP.
% The Numerical Attention module performs ``hard attention,'' copying specific existing variables from earlier positions to the current position based on relational conditions.

\noindent
\textbf{Piecewise Linear Encoding.} 
Since the residual stream consists of raw scalars ($\mathbf{h}_l \in \mathbb{R}^{N_l}$), distinct scalar variables lack the high-dimensional expressivity required for effective dot-product comparisons. To address this, we define the query and key as selected scalars $x_q, x_k \in \mathbb{R}$, and project them into a higher-dimensional space using the Piecewise Linear Encoding (PLE)~\cite{gorishniy2022embeddings}:
\begin{equation}
\mathbf{q} = \phi_{\text{PLE}}(x_q), \quad \mathbf{k} = \phi_{\text{PLE}}(x_k),
\end{equation}
where $\mathbf{q}, \mathbf{k} \in \mathbb{R}^{d_{\text{attn}}}$ and $\phi_{\text{PLE}}: \mathbb{R} \to \mathbb{R}^{d_{\text{attn}}}$ is a learnable mapping.
% This encoding enables the attention mechanism to acquire non-linear decision boundaries---such as activating attention exclusively at the minimum value---which are crucial for implementing algorithmic logic.

% \textbf{Hard Attention.} 
% The attention scores are computed via $A_{raw} = \frac{(Q W_Q)(K W_K)^T}{\sqrt{d_{attn}}} + B_{rel}$, incorporating T5-style relative positional biases \cite{raffel2020exploring}. 
% % Crucially, we enforce a strict ``copy'' semantic by keeping the Value ($v$) projection as an identity mapping for scalars. 
% where the output is a selection of the value $v$ weighted by the discretized attention map:
% \begin{equation}
%     O_{\text{attn}} = S(A_{\text{raw}}, \tau) \cdot v.
% \end{equation}
% By annealing $S(\cdot)$ towards a hard argmax, the module converges to a deterministic pointer operation, retrieving specific values from history (e.g., ``copy the value from the token at offset -1'').
% To ensure interpretable attention patterns, we enforce hard attention using the sampling function $S(\cdot)$. 
% Crucially, the Value ($v$) projection is identity for scalars. The output is a weighted sum of selected scalar values:
% \begin{equation}
%     o_{\text{attn}} = S(A_{\text{raw}}, \tau) \cdot v
% \end{equation}
% This design forces the attention head to converge to a deterministic pointer operation, retrieving specific values from history (e.g., ``copy the value from the token at offset -1''). 

\noindent
\textbf{Deterministic Attention Mechanism.}
The attention scores $\mathbf{a}_{\text{raw}} \in \mathbb{R}^{T}$ over a context length $T$ are computed via scaled dot-product with T5-style relative positional biases $\mathbf{b}_{\text{rel}}$~\cite{raffel2020exploring}:
\begin{equation}
\mathbf{a}_{\text{raw}} = \frac{\mathbf{K} \mathbf{q}}{\sqrt{d_{\text{attn}}}} + \mathbf{b}_{\text{rel}},
\end{equation}
where $\mathbf{K} \in \mathbb{R}^{T \times d_{\text{attn}}}$ stacks the projected keys from the context.
To ensure interpretable attention patterns, we enforce hard attention using the sampling function $S(\cdot, \tau)$. Crucially, the value projection is the identity function for scalars, meaning that the value vector $\mathbf{v} \in \mathbb{R}^T$ consists directly of the raw scalars from the history. The output $o_{\text{attn}} \in \mathbb{R}$ is a weighted sum:
\begin{equation}
o_{\text{attn}} = S(\mathbf{a}_{\text{raw}}, \tau)^\top \mathbf{v}.
\end{equation}
This design forces the attention head to converge to a deterministic pointer operation, retrieving specific raw values from history (e.g., ``copy the value from the token at offset -1'').

\subsection{Numerical MLP as Operator}
\label{subsec:numerical_mlp}

While the Numerical Attention handles data movement, the Numerical MLP is dedicated to element-wise arithmetic and logical transformations.
Drawing on the insight that Transformer MLPs contribute additive updates to the residual stream, which can be decomposed into weighted sums of sub-updates~\cite{geva2022transformer}, we explicitly decompose the MLP module into a collection of parallel, independent sub-modules (sub-MLPs).

% Each sub-module performs a single elementary operation. It first selects a small, fixed number of scalars (typically $k=2$) from the stream via the discretized reading mechanism ($I \in \mathbb{R}^2$). These operands are processed by a shallow network:
% \begin{equation}
%     o_{\text{sub-MLP}} = W_2(\sigma(W_1 I + b_1)) + b_2,
% \end{equation}
% where $\sigma$ is a non-linear activation (e.g., ReLU). The scalar output $y$ is concatenated back to the stream.

Each sub-module performs a single elementary operation. It first selects a small, fixed number of scalars (typically $k=2$\footnote{Unless otherwise specified, we set $k=2$ to capture binary interactions while keeping symbolic regression low-dimensional and stable; see Section~\ref{app:arity_ablation}.}) from the stream via the discretized reading mechanism, forming an operand vector $\mathbf{u} \in \mathbb{R}^k$. These operands are processed by a shallow network:
\begin{equation}
o_{\text{mlp}} = \mathbf{W}_2 \sigma(\mathbf{W}_1 \mathbf{u} + \mathbf{b}_1) + b_2,
\end{equation}
where $\mathbf{W}_1 \in \mathbb{R}^{d_{\text{hid}} \times k}$, $\mathbf{W}_2 \in \mathbb{R}^{1 \times d_{\text{hid}}}$ are learnable weights, $\mathbf{b}_1, b_2$ are biases, and $\sigma$ is a non-linear activation (e.g., ReLU). The resulting scalar output $o_{\text{mlp}}$ is concatenated back to the residual stream.

\noindent
\textbf{Inductive Bias for Symbolic Regression.} By severely constraining the input dimension $k$ and the hidden width $d_{\text{hid}}$, we deliberately limit the complexity of each sub-module. 
This bottleneck forces the network to decompose complex functions into a sequence of simple arithmetic steps, creating ideal conditions for symbolic regression (e.g., PySR~\cite{cranmer2023interpretablemachinelearningscience}) to extract closed-form expressions.

\subsection{Linear Output Head}
\label{subsec:linear_output_head}

% The final component is a linear output head designed to aggregate the discrete computational steps into the final prediction. Given the final residual stream $\mathcal{X}_{final} \in \mathbb{R}^{N_{final} \times 1}$, the output is computed as:
% \begin{equation}
%     \hat{y} = W_{out} \mathcal{X}_{final}
% \end{equation}
% We apply $L_1$ regularization to $W_{out}$ to induce sparsity. This encourages the model to select only the necessary variables from the history of computation, effectively performing feature selection on the generated algorithmic steps. The final trained model thus represents a computational graph where nodes are either copy operations (Attention) or arithmetic functions (MLP), and edges are defined by the discrete selections of the discretized reading modules.

The architecture concludes with a linear output head that aggregates the discrete computational steps into a final prediction:
\begin{equation}
\hat{y} = \mathbf{w}_{\text{out}}^\top \mathbf{h}_{\text{final}},
\end{equation}
where $\mathbf{h}_{\text{final}}$ is the final residual stream, and $\mathbf{w}_{\text{out}}$ represents the aggregation weights. 
% To reduce the length of the extracted program, we impose a sparsity threshold $\epsilon$ on the magnitude of the weights (i.e., set $w_i = 0$ if $|w_i| < \epsilon$). This prunes unnecessary intermediate variables from the computation graph, retaining only the paths essential for the task.
% The architecture concludes with a linear output head that aggregates the discrete computational steps into a final prediction: $\hat{y} = W_{\text{out}} \mathcal{X}_{\text{final}}$.
% To reduce the length of the resulting program, we can impose a threshold on $W_{\text{out}}$ to prune unnecessary intermediate variables while retaining only the computational paths essential for the task.
% To ensure the resulting program is minimal, we apply $L_1$ regularization to $W_{\text{out}}$. This sparsity constraint performs effectively as feature selection, pruning unnecessary intermediate variables and retaining only the computational paths essential for the task. 
% The final trained model thus emerges as a clean, executable program where the computation graph is defined by the non-zero entries in the reading and output matrices.
The final trained model thus represents a clean, executable computational graph: nodes correspond to either routing operations (Numerical Attention) or arithmetic functions (Numerical MLP), and edges are defined by the discrete selections of the discretized reading modules.

\section{From Weights to Code}
\label{sec: from weights to code}

% Once the Discrete Transformer is trained and annealed to a fully discrete state, the learned algorithmic solution is encoded within its weights and activations. To extract an interpretable program, we develop tailored strategies for each architectural component: Numerical Attention, Numerical MLP, and the linear output head. These strategies leverage activations collected during forward passes on a validation dataset, enabling symbolic regression and pattern analysis to recover concise expressions. 
% The extracted elements are then composed into an executable program that faithfully reproduces the model's computation, as validated on algorithmic benchmarks.

After training with temperature annealing, the converged Discrete Transformer reaches a fully discrete state, and the algorithmic solution is implicitly encoded within its sparse weights and activation patterns. 
To recover an explicit, human-readable program, we treat the trained model as a computational graph composed of two distinct types of nodes: \textit{routers} (Attention) and \textit{operators} (MLP). 
% We propose a decoupled extraction pipeline: we first identify the function of each node using component-specific strategies, and then reconstruct the global program trace through backward traversal.
We propose a decoupled extraction pipeline that first infers the function of each node using component-specific strategies, then reconstructs the global program trace via backward traversal.

\subsection{Hypothesis Testing for Numerical Attention}
\label{subsec:attn_extraction}

For the Numerical Attention module, directly extracting explicit symbolic expressions via symbolic regression is challenging due to the computational complexity introduced by the PLE of queries and keys, dot-product interactions, and the hard attention mechanism induced by sampling functions. However, since the structural role of attention is constrained to be an information router, we abstract away from the intermediate arithmetic operations and focus our interpretability analysis on the resulting routing patterns.

We conceptualize the Numerical Attention module as a deterministic pointer performing differentiable addressing on context memory. Following Neural Turing Machines~\cite{graves2014neuralturingmachines}, we categorize addressing into \textit{Location-based} and \textit{Content-based}. We hypothesize that our attention heads specialize into these two modes, and empirically observe that they manifest as two typical interpretable patterns: \textit{Fixed Offset} and \textit{Windowed Extrema}.\footnote{These patterns are not intended as a complete taxonomy of attention behaviors; see Appendix~\ref{app:attention_patterns} for further discussion.}
% We hypothesize and empirically observe that our attention heads specialize into these two modes, manifesting as two typical interpretable patterns: \textit{Fixed Offset} and \textit{Windowed Extrema}, respectively.

% From a theoretical perspective, we conceptualize the converged Numerical Attention module as a deterministic pointer performing \textit{differentiable addressing} on the memory (i.e., the historical context representations). 
% Grounded in the framework of Neural Turing Machines~\cite{graves2014neuralturingmachines}, addressing mechanisms can be fundamentally categorized into two types: \textit{Location-based Addressing}, where access depends on the relative position, and \textit{Content-based Addressing}, where access depends on the value similarity or magnitude. 
% Guided by this taxonomy, we hypothesize that the attention heads in our Discrete Transformer specialize into these two distinct modes. 
% Empirically, across the algorithmic benchmark investigated, we observe that these modes manifest as two typical interpretable patterns: \textit{Fixed Offset} and \textit{Windowed Extrema}.
%
Specifically, for a given head, we analyze its attention matrices generated over the validation set and test the hypotheses by examining the statistical properties:
% Specifically, for a given head, let $\mathbf{A} \in \mathbb{R}^{T \times T}$ denote its attention matrix generated over the validation set, where $A_{ij}$ is the attention weight from query position $i$ to key position $j$.
% We verify the hypotheses by analyzing the statistical behavior of $\mathbf{A}$ across the dataset:
% We characterize each head by computing summary statistics of $\mathbf{A}$ and verifying it against the following templates:
\begin{itemize}
\item \textit{Fixed Offset.}
This pattern corresponds to relative positional indexing.
We hypothesize that there exists an integer offset $\delta$ such that, for a large fraction of query positions $i$, the head places most of its attention on 
$
j = i - \delta
$.
We operationalize this by measuring whether the averaged attention mass concentrates on a single offset diagonal (corresponding to $j=i-\delta$) beyond a predefined threshold.
\item \textit{Windowed Extrema.}
This pattern reflects content-dependent selection.
Let $\mathbf{v} \in \mathbb{R}^T$ be the sequence of scalar values, where $v_j$ is the value at position $j$.
We hypothesize that there exists a window size $z \in \mathbb{Z}^+$ such that, for a large fraction of query positions $i$, the head primarily attends to $j^* = \arg\operatorname*{min/max}_{j \in \{i - z + 1, \dots, i\}} v_j$. 
% i.e., it selects the position within a local backward window whose value is the extremum.
We verify this hypothesis by checking the sample-wise agreement between the head's selected index and the true extremum index within the window.
\end{itemize}
\noindent
There might exist ``unmatched heads''; however, across the algorithm benchmark, we find that they represent computational noise—being negligible in magnitude or acting as redundant variables—and do not influence the logic of the final assembled program (see Appendix~\ref{app:unmatched_heads} for detailed analysis).

\subsection{Symbolic Regression for Numerical MLP}
\label{subsec:mlp_extraction}

In contrast to the Numerical Attention, the Numerical MLP modules serve as the arithmetic core. Thanks to the disentangled architecture, each MLP sub-module functions as an isolated mapping $f: \mathbb{R}^k \to \mathbb{R}$ with low-dimensional inputs. This architectural isolation makes them ideal candidates for black-box symbolic regression.

For each sub-MLP, we collect a dataset of input-output pairs $\mathcal{D} = \{(\mathbf{u}^{(i)}, o^{(i)}_{\text{mlp}})\}_{i=1}^M$ from validation passes. We employ the PySR~\cite{cranmer2023interpretablemachinelearningscience}, a symbolic regression tool based on genetic algorithms, to search for the optimal symbolic expression $\hat{f}$ that minimizes the error on $\mathcal{D}$. Crucially, the constrained input dimension and limited model capacity drastically reduce the search space, enabling the PySR to reliably converge to exact arithmetic expressions (e.g., $o_{\text{mlp}} = 2 u_1 + u_2$) rather than approximate fits.

\subsection{Global Program Assembly}
\label{subsec:program_assembly}

The final phase integrates these extracted primitives into a coherent program. The linear output head, $\hat{y} = \mathbf{w}_{\text{out}}^\top \mathbf{h}_{\text{final}}$, serves as the entry point for extraction.
% We first apply magnitude-based pruning to the output weights, identifying the set of active indices $\mathcal{I} = \{i \mid |w_{\text{out}, i}| > \epsilon\}$ that significantly contribute to the final prediction.
To reduce the length of the extracted program, we impose a sparsity threshold $\epsilon$ on the magnitude of the weights $\mathbf{w}_{\text{out}}$. 
This prunes unnecessary variables from the computation graph, retaining the active variables contributing to the final prediction.
% retaining only the paths essential for the task.
% We first apply magnitude-based pruning to $\mathbf{w}_{\text{out}}$ ($|w_i| > \epsilon$) to identify the active variables contributing to the final prediction.

Starting from these active variables, we perform a \textit{backward traversal} of the computational graph. Recursively, we replace each intermediate variable in the residual stream with its corresponding symbolic definition, either a deterministic pointer from the Numerical Attention or a distilled arithmetic expression from the Numerical MLP, until the traversal reaches the raw input tokens.
This process effectively compiles the neural network into concise, closed-form algorithmic expressions that approximate the Discrete Transformer's behavior with high fidelity across the relevant input domain.

% The final phase integrates these extracted primitives into a coherent program. The linear output head, $\hat{y} = \mathbf{w}_{\text{out}} \mathbf{h}_{\text{final}}$, serves as the entry point for extraction.
% We first apply magnitude-based pruning to $\mathbf{w}_{\text{out}}$ ($|w_i| > \epsilon$) to identify the active variables contributing to the final prediction.

% Starting from these active output variables, we perform a \textit{backward traversal} of the computational graph. Recursively, we replace each intermediate variable in the residual stream with its corresponding symbolic definition—either a verified routing operation or a distilled arithmetic expression—until the traversal reaches the raw input tokens.
% This process effectively compiles the neural network into concise, closed-form algorithmic expressions that provably reproduces the Discrete Transformer's behavior.

\section{Experiments}
\label{sec: Experiments}

In this section, we evaluate the Discrete Transformer on a diverse suite of algorithmic reasoning tasks. We aim to demonstrate that, beyond achieving high predictive performance, our model extracts interpretable and human-readable algorithms with high fidelity, thereby revealing the underlying structure and logic inherent in the data.
% our model not only achieves high performance but also learns interpretable, human-readable algorithms that illuminate the underlying logic of the data.

\noindent
\textbf{Datasets} 
We construct an algorithmic reasoning benchmark designed to evaluate symbolic rule recovery across a range of capabilities, including arithmetic reasoning, variable tracking, and non-linear composition.
The tasks are primarily sourced from the MIPS benchmark \cite{michaud2024openingaiblackbox}, excluding those whose ground-truth rules cannot be expressed using general symbolic operators (e.g., the palindrome rule in \texttt{bit\_palindrome}).\footnote{Since our goal is to extract closed-form expressions of fixed-size through symbolic regression (typically PySR), we restrict our benchmark to tasks whose ground-truth mappings are representable under PySR’s default operator set (e.g., $+, -, \times, \div$).}
% We construct an algorithmic reasoning benchmark designed to evaluate symbolic rule recovery across a range of capabilities, including arithmetic reasoning, variable tracking, and non-linear composition.
% The tasks are primarily sourced from the MIPS benchmark \cite{michaud2024openingaiblackbox}, covering both linear and non-linear logic as well as mathematical and physical rules. We exclude tasks that cannot be expressed as symbolic expressions (e.g., \texttt{bit\_palindrome}).\footnote{Our study focuses on extracting interpretable symbolic expressions via symbolic regression; thus, we restrict our benchmark to tasks whose ground-truth rules are theoretically representable within the target symbolic function space.}
% We evaluate symbolic rule recovery on an algorithmic reasoning benchmark curated from the MIPS suite \cite{michaud2024openingaiblackbox}.
% We construct an algorithmic reasoning benchmark designed to evaluate the capability to learn and recover explicit symbolic rules, ranging from linear arithmetic to non-linear arithmetic. 
% The benchmark is primarily curated from the MIPS suite \cite{michaud2024openingaiblackbox}, selecting tasks whose ground-truth mapping can be expressed as a single-output symbolic function within our expression grammar (e.g., \texttt{sum}), and excluding tasks that require primitives outside our hypothesis class (e.g., \texttt{bit\_palindrome}).\footnote{Our study focuses on extracting closed-form analytical expressions (symbolic regression); thus we restrict to tasks whose ground truth is representable by our symbolic function class.}
To broaden coverage of continuous-valued physical dynamics, we include five continuous-valued tasks: \texttt{exponential\_\allowbreak moving\_\allowbreak average}, \texttt{free\_fall\_height}, \texttt{linear\_drop}, \texttt{quadratic\_drop}, and \texttt{damped\_\allowbreak harmonic\_\allowbreak oscillator}.
We group all tasks into three categories based on their underlying logic: (1) \textit{Linear Arithmetic} (e.g., \texttt{sum\_last2}, \texttt{gravity}), which involves learning linear mathematical or physical logic; (2) \textit{Non-Linear Composition} (e.g., \texttt{parity\_last2}, \texttt{add\_mod\_3}), which requires approximating non-linear logic; and (3) \textit{Continuous Dynamics} (e.g., \texttt{exponential\_moving\_average}), which assess the capability to model tasks with floating-point variables. Detailed task definitions are provided in Appendix Table~\ref{tab:all_experiment_results}.

\noindent
\textbf{Training Details} All models adopt a decoder-only architecture implemented in PyTorch, optimized via AdamW to minimize the mean squared error (MSE). We train for 50 epochs with a batch size of 512 and cosine learning rate decay. Crucially, to handle discrete optimization, we apply geometric annealing to the sampling temperature $\tau$, decreasing it from $10.0$ to $0.1$. We report the average performance across three random seeds, with hyperparameters (layers, heads, sub-MLPs) selected via grid search. Full experimental details are provided in Appendix~\ref{sec: experiment details}.

\noindent
\textbf{Results} 
The Discrete Transformer demonstrates strong predictive performance across most of the algorithmic tasks, with MSE approaching zero (see Appendix Table~\ref{tab:all_experiment_results}).
Beyond prediction, our primary contribution is to show that the solutions encoded in the Discrete Transformer can be \emph{explicitly} extracted into human-readable code.
Specifically, by applying the methodology described in Section \ref{sec: from weights to code}, we successfully distill executable Python programs from the trained weights.
As shown in Table~\ref{tab:experiment_results}, the extracted programs successfully solve most algorithmic tasks, achieving an accuracy of $1.00$, which matches the performance of MIPS.
Moreover, the low root mean squared error (RMSE) further indicates the high fidelity of our extraction methodology.
Notably, the Discrete Transformer additionally handles continuous-dynamics tasks that lie beyond the scope of MIPS.
We summarize three key findings below.

\begin{table*}[t]
    \centering
    \caption{
    Comparison of algorithm extraction performance between MIPS and the Discrete Transformer (Ours) on the algorithmic benchmark. 
    For integer-valued tasks, we evaluate test-set accuracy (Acc.) by rounding outputs to the nearest integer and report Root Mean Squared Error (RMSE) to assess extraction fidelity. 
    For tasks involving floating-point data types, Acc. is not applicable (N/A) and only RMSE is reported.
    MIPS results are retrieved from \citet{michaud2024openingaiblackbox}; $\dagger$ denotes
    that MIPS fails to achieve perfect accuracy ($1.00$) or is inapplicable due to inherent limitations.
    }
    \label{tab:experiment_results}
\begin{small}
\begin{tabular}{@{}lccc lccc@{}}
\toprule
\multirow{2}{*}{\textbf{Task}} & \textbf{MIPS} & \multicolumn{2}{c}{\textbf{Ours}}
& \multirow{2}{*}{\textbf{Task}} & \textbf{MIPS} & \multicolumn{2}{c}{\textbf{Ours}} \\
\cmidrule(lr){2-2} \cmidrule(lr){3-4}
\cmidrule(lr){6-6} \cmidrule(lr){7-8}
& \textbf{Acc.} & \textbf{Acc.} & \textbf{RMSE}
& & \textbf{Acc.} & \textbf{Acc.} & \textbf{RMSE} \\
\midrule

\multicolumn{8}{l}{\textit{Linear Arithmetic}} \\
\texttt{sum\_last2}     & 1.00 & 1.00 & $8.22 \times 10^{-5}$
& \texttt{diff\_last2} & 1.00 & 1.00 & $2.65 \times 10^{-8}$ \\
\texttt{sum}           & 1.00 & 1.00 & $1.30 \times 10^{-7}$
& \texttt{div\_3}           & $\dagger$ & 0.67 & $4.69 \times 10^{-1}$\\
\texttt{freebody}    & 1.00 & 1.00 & $9.78 \times 10^{-6}$ 
& \texttt{gravity}       & 1.00 & 1.00 & $4.31 \times 10^{-6}$\\
\texttt{spring}      & 1.00 & 1.00 & $5.91 \times 10^{-5}$ 
& \texttt{magnetic}      & $\dagger$ & 0.02 & $1.06 \times 10^{1\phantom{-}}$ \\

\midrule
\multicolumn{8}{l}{\textit{Non-Linear Composition}} \\
\texttt{parity\_last2} & 1.00 & 1.00 & $1.21 \times 10^{-6}$
& \texttt{maximum\_prev2}  & 1.00 & 1.00 & $2.10 \times 10^{-3}$ \\
\texttt{minimum\_prev2}    & 1.00 & 1.00 & $4.21 \times 10^{-3}$
& \texttt{unique\_prev2} & 1.00 & 1.00 & $2.05 \times 10^{-7}$ \\
\texttt{bitwise\_and}  & 1.00 & 1.00 & $1.03 \times 10^{-9}$
& \texttt{bitwise\_or} & 1.00 & 1.00 & $1.28 \times 10^{-3}$ \\
\texttt{bitwise\_not}  & 1.00 & 1.00 & $3.80 \times 10^{-9}$
& \texttt{bitwise\_xor} & 1.00 & 1.00 & $2.58 \times 10^{-9}$ \\
\texttt{balanced\_parenthesis} & $\dagger$ & 0.98 & $1.45 \times 10^{-1}$ 
& \texttt{abs}           & 1.00 & 1.00 & $2.29 \times 10^{-7}$ \\
\texttt{abs\_of\_diff} & 1.00 & 1.00 & $1.42 \times 10^{-4}$ 
& \texttt{parity}        & 1.00 & 1.00 & $8.99 \times 10^{-9}$\\
\texttt{parity\_zeros} & 1.00 & 1.00 & $1.99 \times 10^{-5}$ 
& \texttt{add\_mod\_3}   & 1.00 & 1.00 & $4.15 \times 10^{-4}$\\
\texttt{bit\_dot\_prod\_mod\_2} & 1.00 & 1.00 & $1.97 \times 10^{-3}$ 
& \texttt{bit\_addition}   & 1.00 & 0.99 & $1.49 \times 10^{-1}$ \\

\midrule
\multicolumn{8}{l}{\textit{Continuous Dynamics}} \\
\texttt{exponential\_moving\_average} & $\dagger$ & \text{N/A} & $4.04 \times 10^{-7}$
& \texttt{free\_fall\_height} & $\dagger$ & \text{N/A} & $9.02 \times 10^{-3}$ \\
\texttt{linear\_drop} & $\dagger$ & \text{N/A} & $5.21 \times 10^{-4}$ &
\texttt{quadratic\_drop} & $\dagger$ & \text{N/A} & $9.37 \times 10^{-4}$ \\
\texttt{damped\_harmonic\_oscillator} & $\dagger$ & \text{N/A} & $3.18 \times 10^{-4}$ \\

\bottomrule
\end{tabular}

\end{small}
\end{table*}

\begin{itemize}
% \item \textbf{Discovery of Algorithmic Parsimony in Linear Tasks.} In linear tasks, the discrete optimization process demonstrates a strong inductive bias towards \textit{parsimony}. 
% The model frequently learns to bypass the Numerical MLP entirely, leveraging the linear output head to perform arithmetic directly. 
% For instance, in \texttt{sum\_last2} (Figure \ref{fig:sum_last2}), the model assigns specific attention heads to retrieve $x_{t-1}$ (via a verified ``Fixed Offset'' pattern) and integrates it with the current token $x_t$. 
% Symbolic simplification yields the exact expression $y_t = x_t + x_{t-1}$,\footnote{We employ the SymPy library~\cite{meurer2017sympy} to automatically simplify the extracted expressions, leading to more concise and readable mathematical formulae.} effectively capturing the underlying addition logic.
% In physical tasks such as \texttt{gravity}, the Discrete Transformer effectively identifies the essential computational variables and successfully constructs the correct computation graph through the coordination of the Numerical Attention, Numerical MLP, and linear output head, demonstrating strong potential for modeling complex physical processes.

\item \textbf{Discovery of Algorithmic Mechanisms in Linear Tasks.} 
For linear tasks, the discrete optimization process exhibits an efficient ability to discover underlying algorithmic mechanisms. 
On simple instances, the model frequently learns to bypass the Numerical MLP entirely, instead leveraging the linear output head to perform arithmetic directly. 
For instance, in \texttt{sum\_last2} (Figure \ref{fig:sum_last2}),\footnote{For clarity of presentation, we round the coefficients of symbolic expressions to two decimal places.} the model assigns specific attention heads to retrieve $x_{t-1}$ (via a verified ``Fixed Offset'' pattern) and integrates it with the current token $x_t$. 
Symbolic simplification yields the exact expression $y_t = x_t + x_{t-1}$,\footnote{We employ the SymPy library~\cite{meurer2017sympy} to automatically simplify the extracted expressions, leading to more concise and readable mathematical formulae.} effectively capturing the underlying addition logic.
On more complex physical tasks such as \texttt{gravity}, the model identifies the essential computational variables and constructs the correct computation graph through the coordination of the Numerical Attention, Numerical MLP, and linear output head, demonstrating strong potential for efficient modeling of physical processes.

\item \textbf{Exact Recovery of Non-Linear Identities.}
For tasks requiring non-linear logic, the extraction methodology successfully identifies algebraically equivalent expressions rather than merely producing approximate fits.
For instance, in the \texttt{parity\_last2} task ($x_t \oplus x_{t-1}$) with binary inputs $x \in \{0, 1\}$, simplifying the extracted expressions (Figure~\ref{fig:parity_last2}) yields $y_t = x_t + x_{t-1} - 2 x_t x_{t-1}$, which is an exact algebraic formulation of the parity operation.
For \texttt{maximum\_prev2} and \texttt{minimum\_prev2} tasks, simplifying the extracted expressions in Figure~\ref{fig:code_comparison} reveals a piecewise-linear implementation of conditional computation via ReLU operations: $\max(x_t, x_{t-1}) = x_{t-1} + \text{ReLU}(x_t - x_{t-1})$ and $\min(x_t, x_{t-1}) = x_{t-1} - \text{ReLU}(x_{t-1} - x_t)$.
% In the \texttt{parity\_last2} task (XOR operation), utilizing the SymPy library~\cite{meurer2017sympy} for simplification reveals that the model learns the analytical expansion of XOR: $y_t = x_t + x_{t-1} - 2 x_t x_{t-1}$ (see Figure~\ref{fig:parity_last2}).
% Similarly, for \texttt{maximum\_prev2}, the model composes the ReLU activation to implement conditional branching, deriving the identity $\max(a, b) = b + \text{ReLU}(a - b)$ (see Figure \ref{fig:code_comparison}). 
% This demonstrates the model's ability to ``white-box'' standard neural activations into logical control flow.
This demonstrates the capability to reverse-engineer black-box neural computations into explicit algebraic forms equivalent to logical operations.

\item \textbf{Modeling Continuous Dynamics.}
% In physical tasks like \texttt{newton\_gravity}, the Discrete Transformer effectively identifies the necessary computational variables in physical tasks, and successfully implements the correct computation graph through the coordination of the Numerical Attention, Numerical MLP, and the output head, highlighting the Discrete Transformer's potential for interpreting complex reasoning.
% In physical tasks such as \texttt{gravity}, the Discrete Transformer effectively identifies the essential computational variables and successfully constructs the correct computation graph through the coordination of the Numerical Attention, Numerical MLP, and output head, demonstrating strong potential for modeling complex physical processes.
Owing to its inherent model design, our framework offers a distinct qualitative advantage over prior symbolic synthesis approaches in terms of broader applicability.
MIPS, in particular, relies on discrete latent-state abstractions and is therefore not directly applicable to floating-point data types.
% Specifically, MIPS \cite{michaud2024openingaiblackbox} relies on Boolean/integer autoencoders, which inherently limits its scope to discrete data domains and renders it incompatible with floating-point tasks. 
In contrast, the Discrete Transformer natively operates on continuous variables, allowing the extracted programs to preserve real-valued intermediate computations rather than forcing discretization.
% While MIPS \cite{michaud2024openingaiblackbox} relies on Boolean/integer autoencoders—thereby limiting it to discrete data domains and rendering it incapable of handling tasks involving floating-point inputs and outputs—our Discrete Transformer operates natively on continuous floating-point variables. 
This distinction enables the extraction of algorithms that capture continuous dynamics in tasks such as \texttt{exponential\_moving\_average} and \texttt{free\_fall\_height}, substantially broadening the scope of mechanistic-interpretability-based program synthesis.
% For example, in the \texttt{exponential\_moving\_average} and \texttt{free\_fall\_height} tasks, our framework successfully recovers the underlying logic.

\end{itemize}

To evaluate the robustness of our framework under architectural variations, we examine how key hyperparameters, including the number of layers, attention heads, and sub-MLPs, affect both convergence (MSE) and program complexity (measured by the line count). 
Across the tested tasks and architectural configurations, our results indicate that once the architecture meets the minimal functional requirements of the task, the Discrete Transformer generally attains near-zero MSE loss and yields executable extracted programs.

Over-parameterized architectures remain robust in terms of task performance, but they can lead to increased redundancy in the extracted programs. In particular, multiple heads or sub-MLPs may learn functionally similar operations, increasing the program line count without yielding commensurate gains in accuracy. We find that such redundancy can be substantially reduced by imposing a mild simplicity bias, such as $\ell_1$ regularization on the linear output head, while maintaining low extraction error. Moreover, excess capacity may enlarge the discrete search space and make optimization less stable, but a slower temperature annealing schedule alleviates this effect. Overall, these results indicate that the Discrete Transformer is robust across a range of architectural choices: once sufficient functional capacity is available, it reliably recovers accurate discrete solutions, while the additional complexity introduced by excess capacity can be controlled through regularization and annealing. Additional details are provided in Appendix~\ref{app:robustness}.

% \paragraph{Linear Tasks (State-Tracking \& Non-State-Tracking).}
% In these tasks, the discrete optimization process effectively guides the model to utilize Numerical Attention strictly for information routing. Interestingly, for elementary arithmetic, the model frequently bypasses the Numerical MLP, directly leveraging the linear output head to execute operations. 
% As illustrated in Figure \ref{fig:sum_last2}, for the \texttt{sum\_last2} task, the Discrete Transformer learns to attend to the previous position ($t-1$) using Numerical Attention. The linear output head then aggregates the current input $x_t$ and the retrieved $x_{t-1}$ with both weights approximating $1.0$, effectively implementing the addition logic.

\begin{figure}[t]
\centering
\begin{minipage}{\columnwidth}
\begin{lstlisting}
import numpy as np
def sum_last2(input_seq):
    # V0_input
    input_arr = np.array(input_seq, dtype=float)
    seq_len = input_arr.shape[0]
    V0 = input_arr[:, 0]
    # V1_Attn_L0H0
    V1 = np.zeros(seq_len)
    # Fixed offset 1
    V1[1:] = V0[:-1]
    # V2_Attn_L0H1
    V2 = np.zeros(seq_len)
    # Fixed offset 1
    V2[1:] = V0[:-1]
    # Output Head
    output = 1.00 * V0 + 1.00 * V1 + 0.01 * V2
    return output
\end{lstlisting}
% --- 分割线 ---
% \vspace{-0.6em}
\begin{center}
\small
$\Downarrow$ \; \textit{symbolic simplification}
\end{center}
% \vspace{-0.7em}
\begin{tcolorbox}[
  colback=formbg,
  frame hidden,
  boxrule=0pt,
  left=6pt,right=6pt,top=2pt,bottom=2pt,
  fontupper=\small,
]
\textbf{Simplified Expression:}\quad
$\displaystyle y_t = x_t + x_{t-1}$
\end{tcolorbox}
% \begin{lstlisting}
% **
% \textbf{Simplified expression:}\quad
% y_t = x_t + x_{t-1}
% **
% \end{lstlisting}
% \vspace{-1em} % 调整间距
% \hrulefill
% \vspace{0.5em}
% % --- 公式化简部分 ---
% \centering
% \small
% \textbf{Formula:} $y_t = x_t + x_{t-1}$ 
% 这里根据代码逻辑：V0是x_t, V1/V2是x_{t-1}，系数和为1
% \vspace{0.3em}
% \small
% \textit{(Simplification via SymPy \cite{meurer2017sympy}: constant folding and term collection)}
\end{minipage}
% \vspace{-2mm}
\caption{
Algorithm extraction results for the \texttt{sum\_last2} task.
Modules are denoted by their type and indices (e.g., \texttt{Attn\_L0H0} represents the attention head at index 0 of layer 0).
% Throughout the paper, modules are denoted by their type and indices (e.g., \texttt{Attn\_L0H0} denotes the attention head at index 0 of layer 0).
The extracted code reveals that the model utilizes specific attention heads to retrieve the previous token. 
Symbolic simplification (bottom) shows the mathematically simplified expression, verifying that the model correctly learns the logic $y_t = x_t + x_{t-1}$.
}
\label{fig:sum_last2}
\end{figure}

\begin{figure}[t]
\centering
\begin{minipage}{\columnwidth}
\begin{lstlisting}
import numpy as np
def parity_last2(input_seq):
    # V0_input
    input_arr = np.array(input_seq, dtype=float)
    seq_len = input_arr.shape[0]
    V0 = input_arr[:, 0]
    # V1_Attn_L0H0
    V1 = np.zeros(seq_len)
    # Fixed offset 1
    V1[1:] = V0[:-1]
    # V2_Attn_L0H1
    V2 = np.zeros(seq_len)
    # Fixed offset 1
    V2[1:] = V0[:-1]
    # V3_MLP_L0M0
    V3 = ((V1 + -0.41) * ((V0 - 0.41) * 3.57)) + -0.61
    # Output Head
    output = -0.56 * V3 + 0.17 * V0 + 0.10 * V2 + 0.07 * V1
    return output
\end{lstlisting}
% --- 分割线 ---
% \vspace{-0.6em}
\begin{center}
\small
$\Downarrow$ \; \textit{symbolic simplification}
\end{center}
% \vspace{-0.7em}
\begin{tcolorbox}[
  colback=formbg,
  frame hidden,
  boxrule=0pt,
  left=6pt,right=6pt,top=2pt,bottom=2pt,
  fontupper=\small,
]
\textbf{Simplified Expression:}\quad
$\displaystyle y_t = x_t + x_{t-1} - 2 x_t x_{t-1}$
\end{tcolorbox}
% \vspace{-1em} % 调整间距
% \hrulefill
% \vspace{0.5em}
% % --- 公式化简部分 ---
% \centering
% \small
% \textbf{Formula:} $y_t = x_t + x_{t-1} - 2 x_t x_{t-1}$
\end{minipage}
% \vspace{-2mm}
\caption{
Algorithm extraction results for the \texttt{parity\_last2} task.
The extracted code reveals that the model utilizes specific attention heads (e.g., \texttt{Attn\_L0H0}) to retrieve the previous token, and specific sub-MLPs (e.g., \texttt{MLP\_L0M0}) to perform non-linear transformations.
The bottom panel presents the symbolic simplification $y_t = x_t + x_{t-1} - 2 x_t x_{t-1}$, which is the algebraic formulation of the parity logic.
}
\label{fig:parity_last2}
\end{figure}

\begin{figure}[t] 
\centering
\begin{minipage}[t]{\columnwidth}
\begin{lstlisting}
import numpy as np
def relu(x):
    return np.maximum(0, x)
def extrema_prev2(input_seq, mode):
    # V0_input
    input_arr = np.array(input_seq, dtype=float)
    seq_len = input_arr.shape[0]
    V0 = input_arr[:, 0]
    # V1_Attn_L0H0
    V1 = np.zeros(seq_len)
    # Fixed offset 1
    V1[1:] = V0[:-1]
    # V3_MLP_L0M0, Output Head
    if mode == "max":  # maximum_prev2
        V3 = (((V0 * -0.12) + relu((V1 * -1.00) + V0)) * -2.21) + ((V1 * -0.26) / 1.02)
        output = 0.88 * V1 + -0.45 * V3 + 0.12 * V0
    else:  # minimum_prev2
        V3 = 1.43 * (V0 + (((3.49 * relu(V1 - V0)) - V1) - (V1 * 1.04)))
        output = 0.42 * V1 + 0.29 * V0 + -0.20 * V3
    return output
\end{lstlisting}
% \vspace{-0.6em}
\begin{center}
\small
$\Downarrow$ \; \textit{symbolic simplification}
\end{center}
% \vspace{-0.7em}
\begin{tcolorbox}[
  colback=formbg,
  frame hidden,
  boxrule=0pt,
  left=6pt,right=6pt,top=2pt,bottom=2pt,
  fontupper=\small,
]
\textbf{Simplified Expressions:}
\[
\begin{aligned}
\max(x_t, x_{t-1})
&= x_{t-1} + \text{ReLU}(x_t - x_{t-1})\\
\min(x_t, x_{t-1})
&= x_{t-1} - \text{ReLU}(x_{t-1} - x_t)
\end{aligned}
\]
\end{tcolorbox}
\end{minipage}

% \begin{minipage}[t]{\columnwidth}
% \begin{lstlisting}
% import numpy as np
% def relu(x):
%     return np.maximum(0, x)
% def minimum_prev2(input_seq):
%     # V0_input
%     input_arr = np.array(input_seq, dtype=float)
%     seq_len = input_arr.shape[0]
%     V0= input_arr[:, 0]
%     # V1_Attn_L0H0 (attention)
%     V1 = np.zeros(seq_len)
%     # Fixed offset 1
%     if seq_len > 1:
%         V1[1:] = V0[:-1]
%     # V3_MLP_L0M0 (mlp)
%     V3 = 1.43 * (V0 + (((3.49 * relu(V1 - V0)) - V1) - (V1 * 1.04)))
%     # Output Head
%     output = 0.42 * V1 + 0.29 * V0 + -0.20 * V3
%     return output
% \end{lstlisting}
% \vspace{-0.6em}
% \begin{center}
% \small
% $\Downarrow$ \; \textit{symbolic simplification}
% \end{center}
% \vspace{-0.7em}
% \begin{tcolorbox}[
%   colback=formbg,
%   frame hidden,
%   boxrule=0pt,
%   left=6pt,right=6pt,top=2pt,bottom=2pt,
%   fontupper=\small,
% ]
% \textbf{Simplified Expression:}\quad
% $\displaystyle \min(x_t, x_{t-1}) = x_{t-1} - \text{ReLU}(x_{t-1} - x_t)$
% \end{tcolorbox}
% \end{minipage}
% \vspace{-2mm} % 调整 caption 与代码块的垂直距离
\caption{
Algorithm extraction results for \texttt{maximum\_prev2} and \texttt{minimum\_prev2} tasks.
The top panel shows the raw code where sub-MLPs utilize ReLU functions to compare the current token $x_t$ with the previous token $x_{t-1}$.
The bottom panel presents the simplified expressions, verifying that the model correctly reconstructs the extrema functions using the ReLU-based algebraic identities.
}
\label{fig:code_comparison}
\end{figure}

\section{Further Analysis}
\label{sec: Further Analysis}

In this section, we provide empirical analyses of the Discrete Transformer’s properties, positioning it at the intersection of program synthesis and continuous sparsification. 
We primarily focus on the training dynamics, characterizing a distinct phase where functional convergence is achieved prior to complete structural discretization.
We further demonstrate how architectural constraints serve as strong inductive biases, establishing the model as a controllable testbed for interpretable algorithm discovery.

\begin{figure}[t]
    \centering
    \includegraphics[width=\columnwidth]{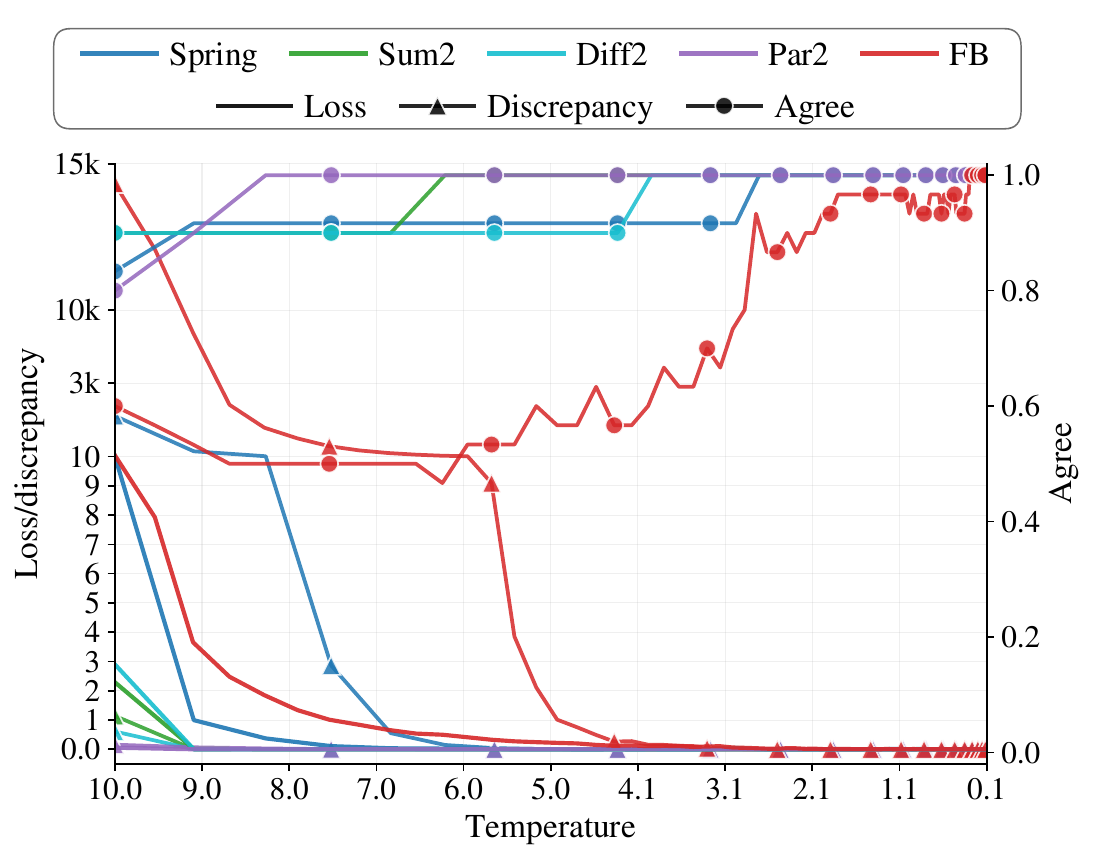}
    \caption{
    Training dynamics exhibit a distinct phase transition: the loss decreases earlier, while the pronounced drop in Discrepancy occurs slightly later, particularly for more challenging tasks such as \texttt{spring} and \texttt{freebody}, coinciding with Agreement approaching $1.0$ during temperature annealing from $10.0$ to $0.1$.
    The abbreviations Spring, Sum2, Diff2, Par2, and FB denote the \texttt{spring}, \texttt{sum\_last2}, \texttt{diff\_last2}, \texttt{parity\_last2}, and \texttt{freebody} tasks, respectively.
    }
    \label{fig:dynamics}
\end{figure}

\subsection{Continuous-to-Discrete Homotopy}

We frame program synthesis as a continuous-to-discrete homotopic transformation.
From the perspective of differentiable architecture search and continuous sparsification \cite{louizos2017learning, jang2016categorical}, we relax the discrete constraint by introducing continuous structural parameters, specifically the projection matrices $\mathbf{W}_{\text{read}}$.
% we relax the combinatorial optimization problem by introducing continuous structural parameters $\mathbf{W}_{\text{read}}$.
The temperature $\tau$ serves as a homotopy parameter: high values define a search space over the continuous probability simplex for exploration, while annealing $\tau \to 0$ smoothly deforms the distribution toward simplex vertices for exploitation.

To validate these dynamics, in addition to the MSE loss (serving as the soft training objective $\mathcal{L}_{\text{soft}}$), we monitor two metrics: Structural Agreement ($\mathcal{A}(e)$) and Discretization Discrepancy ($\Delta(e)$).
Let $\{\mathbf{p}_r^{(e)}\}_{r=1}^R$ be the set of all row-wise probability distributions derived from $S(\mathbf{W}_{\text{read}}, \tau)$ at epoch $e$, $E$ the total number of epochs, and $\mathcal{L}_{\text{hard}}(e)$ the hard loss computed via deterministic argmax sampling.
We define $\mathcal{A}(e) = \frac{1}{R}\sum_{r=1}^{R}
\mathbb{I}\!\left(\arg\max \mathbf{p}_r^{(e)} = \arg\max \mathbf{p}_r^{(E)}\right)$.
And we further define $\Delta(e) = \mathcal{L}_{\text{hard}}(e) - \mathcal{L}_{\text{soft}}(e)$.
% We define:
% *\mathcal{A}(e) = \frac{1}{R}\sum_{r=1}^{R} \mathbb{I}\left(\arg\max \mathbf{p}_r^{(e)} = \arg\max \mathbf{p}_r^{(E)}\right), \quad \Delta(e) = \mathcal{L}_{\text{hard}}(e) - \mathcal{L}_{\text{soft}}(e).*
As shown in Figure~\ref{fig:dynamics}, we observe a distinct phase transition: the significant decline in $\Delta(e)$ occurs slightly later than that of $\mathcal{L}_{\text{soft}}$, concurrently with $\mathcal{A}(e)$ approaching $1.0$ as annealing proceeds.
This lag suggests a two-stage process where the model first achieves \textit{soft convergence}—learning functional mappings via relaxed representations—before undergoing \textit{structural crystallization}, thereby ensuring a robust transition from exploration to exploitation.

\subsection{Controllability via Inductive Biases} \label{subsec:controllability}

Unlike standard code LLMs, which synthesize programs based on opaque statistical patterns, our Discrete Transformer offers a unique advantage: interpretability-aware controllability.
% In scenarios like algorithm extraction, it is often necessary to explore specific regions of the solution space or enforce particular inductive biases.
In scientific discovery, researchers often seek not just any solution, but a specific form of algorithm that aligns with domain knowledge \cite{schmidt2009distilling, udrescu2020ai, cranmer2020discoveringsymbolicmodelsdeep}. 
We address this need by explicitly manipulating the architectural constraints and training configurations of our model to impose inductive biases, thereby steering the solution space.

We demonstrate this controllability through two intervention scenarios.
First, we manipulate architectural primitives. 
In the \texttt{maximum\_prev2} task, the unconstrained model typically solves maximization via MLP-based arithmetic approximation (exploiting ReLU non-linearity).
To test if the model can switch algorithmic paradigms, we explicitly set the number of sub-MLPs to zero, removing its capacity for complex arithmetic.
Consequently, the model adapts by shifting its entire mechanism to the Numerical Attention module. It discovers a ``windowed max'' attention pattern, solving the task by directly copying the largest value from the context rather than computing it.
% However, by defining the problem as ``comparison-based and disabling the sub-MLP modules, we force the model to rely solely on the Numerical Attention mechanism. Consequently, the model successfully discovers a ``windowed max'' attention pattern to select the maximum value directly.
Second, we intervene in the information flow. In the \texttt{spring} task, the model typically learns the standard recurrence $y_t = y_{t-1} - y_{t-2} + x_t$. By masking the immediate history ($y_{t-1}, y_{t-2}$), we guide the model to bypass the standard path and discover a mathematically equivalent high-order recurrence: $y_t = -y_{t-3} + x_t + x_{t-1}$.
These findings highlight the Discrete Transformer as a robust tool for intervenable algorithm discovery, capable of uncovering multiple equivalent logical paths underlying the same data distribution.

\section{Limitations}
\label{sec:limitations}

The Discrete Transformer intentionally trades expressivity for interpretability. Its router-style attention is designed mainly for explicit information routing, and its sub-MLPs operate with bounded fan-in. Consequently, an $L$-layer model can be viewed as a selector-augmented bounded-depth computation graph, where information from many positions is aggregated progressively across layers. Computations requiring broad aggregation or state propagation may therefore require depth that grows with the sequence length. Our experiments are conducted at benchmark scale with relatively small models, and are intended to establish a controlled framework for high-fidelity program extraction. Extending this approach to larger models, richer program classes, and more compositionally complex tasks remains an important direction for future work.

\section{Conclusion}
\label{sec:conclusion}

In this work, we present the Discrete Transformer, a novel architecture enabling program synthesis via algorithm extraction directly from Transformer architectures. 
By combining a disentangled numerical residual stream with a smooth discrete optimization curriculum, the model separates information routing from arithmetic computation and converges to a discrete, extractable representation. We then recover the underlying algorithm through a modular extraction procedure, using hypothesis testing to identify interpretable routing patterns and symbolic regression to derive the corresponding arithmetic expressions.
Across a range of algorithmic benchmarks, our experiments show that this framework achieves performance comparable to RNN-based approaches, while offering fine-grained controllability and significantly broadening the applicability of mechanistic-interpretability-based synthesis.

More broadly, our results suggest that architectural constraints can make Transformer computation more transparent while preserving the ability to solve nontrivial algorithmic tasks. Rather than treating interpretability as a post-hoc analysis of opaque models, the Discrete Transformer embeds interpretability directly into both the architectural design and optimization process. We hope this work provides a step toward faithful algorithm extraction from neural models and motivates future research on bridging interpretability with richer compositional expressivity.

\section*{Acknowledgments}

This research is supported by the National Natural Science Foundation of China under Grant Nos. 62192731.

\section*{Impact Statement}

This paper advances the fields of mechanistic interpretability and algorithm extraction by enabling the discovery of verifiable, human-readable programs from the Discrete Transformer. There are many potential societal consequences of our work, none of which we feel must be specifically highlighted here.


\bibliography{example_paper}
\bibliographystyle{icml2026}

%%%%%%%%%%%%%%%%%%%%%%%%%%%%%%%%%%%%%%%%%%%%%%%%%%%%%%%%%%%%%%%%%%%%%%%%%%%%%%%
%%%%%%%%%%%%%%%%%%%%%%%%%%%%%%%%%%%%%%%%%%%%%%%%%%%%%%%%%%%%%%%%%%%%%%%%%%%%%%%
% APPENDIX
%%%%%%%%%%%%%%%%%%%%%%%%%%%%%%%%%%%%%%%%%%%%%%%%%%%%%%%%%%%%%%%%%%%%%%%%%%%%%%%
%%%%%%%%%%%%%%%%%%%%%%%%%%%%%%%%%%%%%%%%%%%%%%%%%%%%%%%%%%%%%%%%%%%%%%%%%%%%%%%
\newpage
\appendix
\onecolumn
\section{Smooth Transition Mechanism for Discrete Optimization}
\label{sec: smooth transition mechanism}

% To train the discrete selection parameters in our Discretized Reading and Numerical Attention modules, we address the non-differentiability of hard selection while avoiding the pitfalls of standard relaxation methods. While Gumbel-Softmax enables differentiable sampling, it often struggles to escape local optima and fails to promote the strict sparsity essential for interpretability.To address this, we incorporate a smooth transition mechanism \cite{ICLR2025_9d560961}, which dynamically interpolates between Gumbel-Softmax and Gumbel-Sparsemax \cite{martins2016softmax} throughout the training process. The hybrid sampling vector $\mathbf{s} \in \mathbb{R}^{N_l}$ is defined as:\begin{equation}\mathbf{s} = (1-\alpha(\tau)) \mathbf{s}{\text{soft}} + \alpha(\tau) \mathbf{s}{\text{sparse}},\end{equation}where $\mathbf{s}_{\text{soft}}$ and $\mathbf{s}_{\text{sparse}}$ denote sample vectors drawn from Gumbel-Softmax and Gumbel-Sparsemax distributions, respectively. The interpolation coefficient $\alpha(\tau) \in [0, 1]$ is a scheduler function of the temperature $\tau$, designed to shift strictly from 0 to 1 as $\tau$ anneals \cite{ICLR2025_9d560961}.This mechanism allows the model to prioritize exploration via Softmax in the early stages, before smoothly transitioning to Sparsemax to enforce sparsity and deterministic selection (exploitation) in later stages. Ultimately, this ensures convergence to concise, interpretable program structures.

% \textbf{Smooth Transition Mechanism for Discrete Optimization.} 
To train the discrete selection parameters in our discretized reading and Numerical Attention modules, we address the non-differentiability of hard selection while avoiding the pitfalls of standard relaxation methods. 
While Gumbel-Softmax enables differentiable sampling, it often struggles to escape local optima and fails to promote the strict sparsity essential for interpretability. 
To address this, we incorporate a smooth transition mechanism \cite{ICLR2025_9d560961}, which dynamically interpolates between Gumbel-Softmax and Sparsemax \cite{martins2016softmax} throughout the training process. 
The hybrid sampling vector $\mathbf{p} \in \mathbb{R}^{N_l}$ is defined as:
\begin{equation}
\mathbf{p} = (1-\alpha(\tau)) \mathbf{p}_{\text{soft}} + \alpha(\tau) \mathbf{p}_{\text{sparse}},\end{equation}
where $\mathbf{p}_{\text{soft}}$ and $\mathbf{p}_{\text{sparse}}$ denote sample vectors drawn from Gumbel-Softmax and Gumbel-Sparsemax distributions, respectively. 
% The interpolation factor $\alpha(\tau)$ is a function of the temperature $\tau$ ($\alpha(\tau) = \frac{\tau_1 - \tau}{\tau_1 - \tau_2}$), designed to shift strictly from 0 to 1 as $\tau$ geometrically anneals.
The interpolation factor $\alpha(\tau)$ is a temperature-dependent function, defined as $\alpha(\tau) = \frac{\tau_1 - \tau}{\tau_1 - \tau_2}$ (where $\tau_1$ denotes the initial temperature and $\tau_2$ denotes the final temperature), which monotonically transitions from $0$ to $1$ as the temperature $\tau$ is geometrically annealed.
% The interpolation coefficient $\alpha(\tau) \in [0, 1]$ is a scheduler function (typically geometric annealing) of the temperature $\tau$, designed to shift strictly from $0$ to $1$ as $\tau$ anneals.
% The interpolation coefficient $\alpha(\tau) \in [0, 1]$ is a temperature-dependent scheduling function (typically implemented via geometric annealing) that monotonically transitions from $0$ to $1$ as the temperature $\tau$ is annealed.
This mechanism allows the model to prioritize exploration via Softmax in the early stages, before smoothly transitioning to Sparsemax to enforce sparsity and deterministic selection (exploitation) in later stages. 
Ultimately, this ensures convergence to concise, interpretable program structures.

% The hybrid sampling output $y$ is defined as:
% \begin{equation}
%     y = (1-\alpha(\tau)) \cdot y_{soft} + \alpha(\tau) \cdot y_{sparse}
% \end{equation}
% where $y_{soft}$ and $y_{sparse}$ denote samples drawn from Gumbel-Softmax and Gumbel-Sparsemax, respectively. The interpolation factor $\alpha(\tau)$ is a function of the temperature $\tau$, designed to shift strictly from 0 to 1 as $\tau$ anneals. 
% The mechanism allows the model to prioritize exploration via Softmax in the early stages of training, before smoothly transitioning to Sparsemax to enforce sparsity and deterministic selection (exploitation) in later stages. 
% Ultimately, this ensures the convergence to concise, interpretable program structures.

\section{Arity Ablation}
\label{app:arity_ablation}

We use \(k=2\) by default because it is the smallest arity that captures many binary arithmetic primitives while keeping symbolic regression tractable.

Ablations show the expected trade-off.
For a small model with Layers \(=2\), Heads \(=2\), and sub-MLPs \(=2\), \(k=1\) underfits interaction-heavy tasks such as \texttt{add\_mod\_3} and \texttt{bit\_dot\_prod\_mod\_2}, yielding RMSE \(6.46\times10^{-2}\) and \(2.36\times10^{-1}\), whereas \(k=2\) reduces them to \(3.10\times10^{-4}\) and \(3.00\times10^{-4}\). Increasing arity further does not uniformly help: \(k=4\) gives \(2.98\times10^{-3}\) and \(8.44\times10^{-4}\) on the same tasks, indicating a larger symbolic regression search space. For \texttt{parity\_last3}, which requires a genuine three-way interaction, \(k=3\) performs best with RMSE \(3.33\times10^{-6}\), while \(k=2\) fails \((4.69\times10^{-1})\) and \(k=4\) also succeeds but less cleanly \((2.06\times10^{-3})\). The same trend appears in a larger model \((L=4,H=4,M=4)\): increasing \(k\) from 2 to 4 worsens \texttt{add\_mod\_3} and \texttt{bit\_dot\_prod\_mod\_2}, while only modestly improving \texttt{parity\_last3}.

\section{Unmatched Heads}
\label{app:unmatched_heads}

% While the \textit{Fixed Offset} and \textit{Windowed Extrema} patterns achieve high coverage across the algorithmic benchmark investigated, we observe a subset of attention heads that do not conform to these interpretability templates.
Our analysis reveals that the ``unmatched heads'' do not represent missing algorithmic primitives, but rather manifest as computational noise or redundancy.
Specifically, we observe that these heads are generally rendered ineffective through two primary mechanisms: by being suppressed with negligible magnitudes, or by being ignored by higher-layer modules.
% These heads typically fall into two categories: (1) \textit{Magnitude-Pruned Heads}, which exhibit negligible output weights; and (2) \textit{Functionally Redundant Heads}, which generate variables that become ``dead code'' in the final computational graph.  
% In software engineering terms, these heads act as ``dead code'' that can be safely eliminated without altering the program's logic.

For instance, in the \texttt{bitwise\_and} task, the head labeled \texttt{Attn\_L0H1} exhibits a disordered attention matrix that matches neither location-based nor content-based patterns.
However, symbolic tracing of the computational graph confirms that its output variable is an orphan variable: it is not selected as an input by any subsequent modules nor aggregated by the final output projection.
Similarly, in the \texttt{minimum\_prev2} task with the number of sub-MLPs constrained to zero, we observe that the head labeled \texttt{Attn\_L0H1} behaves as an unmatched head. Since its output is disconnected from the valid execution path, this lack of interpretability does not impede successful extraction of the underlying algorithm.
% Similarly, in the \texttt{minimum\_prev2} task constrained with number of sub-MLPs zero, we observe that the head labeled \texttt{Attn\_L0H1} act as an unmatched head.
% Since their outputs are disconnected from the valid execution path, their non-interpretability does not hinder the successful extraction of the underlying algorithm.

\section{Attention Patterns}
\label{app:attention_patterns}

The attention patterns considered in Section~\ref{subsec:attn_extraction}, namely \textit{Fixed Offset} and \textit{Windowed Extrema}, are not intended to constitute a complete taxonomy of attention behaviors. Instead, they represent the dominant and typically interpretable head classes induced by our architecture and benchmark suite.

For tasks in which attention naturally performs computations beyond pure routing, the set of hypothesized attention classes can be extended without changing the overall extraction pipeline. As diagnostic examples, we consider two additional tasks: \texttt{retrieve\_value\_by\_key} and \texttt{prefix\_mean}. In \texttt{retrieve\_value\_by\_key}, we introduce an equality-retrieval head class, which selects values according to key equality rather than relative position or local extrema. In \texttt{prefix\_mean}, we retain soft attention and regard mean aggregation as an additional computational operator.
These extensions preserve the structure of the backward reconstruction procedure, yielding extracted programs with RMSEs of $1.28 \times 10^{-10}$ on \texttt{retrieve\_value\_by\_key} and $3.14 \times 10^{-4}$ on \texttt{prefix\_mean}, respectively, thereby demonstrating that the framework can accommodate additional interpretable attention mechanisms beyond the two default patterns.

\section{Magnitude-based Pruning}
\label{app:magnitude-based_pruning}

To reduce the length of the extracted program, we apply a magnitude threshold \(\epsilon\) to the output weights.
This is a heuristic pruning step: it removes variables that are not needed by the reconstructed dependency graph under the chosen threshold, and we empirically verify that it preserves predictive fidelity on representative tasks (see Table~\ref{tab:pruning_ablation}).

\begin{table}[t]
\centering
\caption{Effect of output-head pruning on extraction fidelity and program length. Values in parentheses are without pruning.}
\label{tab:pruning_ablation}
\begin{tabular}{lcc}
\toprule
Task & RMSE & Program length \\
\midrule
\texttt{sum\_last2} & $1.05\times10^{-4}$ ($1.05\times10^{-4}$) & 28 (34) \\
\texttt{diff\_last2} & $0$ ($2.00\times10^{-8}$) & 22 (33) \\
\texttt{parity\_last2} & $1.49\times10^{-8}$ ($1.49\times10^{-8}$) & 34 (34) \\
\bottomrule
\end{tabular}
\end{table}

\section{Experiment Details}
\label{sec: experiment details}

\noindent
\textbf{Datasets}
The investigated tasks probe specific capabilities of neural networks, such as arithmetic reasoning, variable tracking, and non-linear composition. We categorize them into three logical tiers based on their underlying complexity:
\begin{itemize}
    \item \textbf{Linear Arithmetic.} 
    This category involves linear transformations. \textit{Non-state-tracking} tasks (e.g., \texttt{sum\_last2}) require only local attention within a fixed window. In contrast, \textit{State-tracking} tasks (e.g., \texttt{sum\_last}) require the model to maintain a persistent memory state—implicitly simulating a finite state automaton \cite{zhang2025finitestateautomatainside}—to compute the recurrent state over the sequence.
    Moreover, to evaluate generalization beyond pure arithmetic, we include tasks derived from classical mechanics (\texttt{freebody}, \texttt{gravity}, and \texttt{spring}), which are mathematically reducible to linear state-tracking tasks.
    
    \item \textbf{Non-Linear Composition.} 
    These tasks necessitate non-linear activation logic. Variants range from local operations like \texttt{parity\_last2} and \texttt{maximum\_prev2} to global state-tracking tasks like \texttt{add\_mod\_3}. Success here requires the Numerical MLP to approximate non-linear functions (e.g., XOR) rather than simple linear aggregation.
    
    \item \textbf{Continuous Dynamics.} 
    While the preceding two categories primarily focus on discrete variables, this category encompasses mathematical and physical tasks that require modeling and processing continuous-valued variables, such as \texttt{exponential\_moving\_average} and \texttt{free\_fall\_height}.
    % The above two categories mainly focus on discrete variables, while these are some mathematical or physical tasks requiring precessing continuous-valued variables, such as \texttt{exponential\_moving\_average} and \texttt{free\_fall\_height}. 
    
    % To evaluate generalization beyond pure arithmetic, we include tasks derived from classical mechanics (\texttt{freebody}, \texttt{gravity}, and \texttt{spring}). While mathematically reducible to iterative linear updates, these tasks challenge the model to discover governing physical laws (e.g., Newton's laws of motion) and simulate continuous dynamical systems purely from observed data.
\end{itemize}

\noindent
\textbf{Piecewise Linear Encoding.}
For Numerical Attention, scalar variables selected by the constrained read matrices are first lifted to a higher-dimensional representation using Piecewise Linear Encoding (PLE). Specifically, we partition the numerical range of the training data into bins and learn an embedding vector for each bin boundary. A scalar value is then represented by linearly interpolating between the embeddings of its two neighboring boundaries.

\noindent
\textbf{T5-style Relative Positional Biases.}
In addition to content-based scores, each Numerical Attention head uses a T5-style relative positional bias. For every query-key pair, we compute the clipped relative offset between their sequence positions and map this offset to a learnable head-specific bias, which is added to the attention score before causal masking and sparse sampling.

\noindent
\textbf{Training Details}
All models adopt a decoder-only architecture implemented in PyTorch \cite{paszke2019pytorchimperativestylehighperformance}, optimized via AdamW \cite{loshchilov2017decoupled} to minimize MSE. The training dataset consists of $1,000,000$ samples. Unless otherwise specified, both the input and output sequences have a fixed length of $10$. We employ a cosine annealing schedule to decay the learning rate from $0.05$ to $1 \times 10^{-6}$ over 50 epochs with a batch size of 512. (For the \texttt{freebody} and \texttt{gravity} tasks, models are trained for 100 epochs with an initial learning rate of 0.01.) To handle discrete optimization, we apply a geometric annealing schedule to the sampling temperature $\tau$, decreasing it from $10.0$ to $0.1$ to facilitate a gradual transition from continuous exploration to discrete selection.
For each task, we conduct a grid search over the number of layers $\{1, 2\}$, attention heads $\{2, 3\}$, and sub-MLPs per layer $\{2, 3\}$. We run all experiments across three random seeds, reporting the mean performance and selecting the checkpoint with the best validation performance for subsequent analysis.

\noindent
\textbf{Task Formulation}
% \label{subsec:task_formulation}
% To ensure that the learned modules capture consistent functions—a prerequisite for successful symbolic regression—we categorize algorithmic tasks into two classes based on their computational structure and tailor the training paradigm accordingly.
% To facilitate algorithm extraction and promote both length extrapolation and positional robustness, we aim for the learned modules to capture consistent functions invariant to sequence length and absolute position. To this end, we categorize tasks based on the presence of state-tracking and tailor the task formulation to encourage this consistency. 
% Specifically, for the non-state-tracking tasks, we formulate them as Token-Tagging problems, where the model needs to predict an output across every position.
% In contrast, for the state-tracking tasks, we formulate them as Language Modeling problems, where the model needs to do next-token prediction based on input and previous computation.
% To facilitate symbolic regression and align with the regression nature of the task, we adopt Mean Squared Error (MSE) as the loss function, enabling the model to learn continuous functional mappings.
To facilitate algorithm extraction across varying positions, we design our task formulations to encourage the learning of position-invariant functions. Specifically, we categorize tasks according to their reliance on state-tracking: (1) Non-state-tracking tasks are formulated as Token-Tagging problems, requiring independent, element-wise predictions for each position. (2) State-tracking tasks are framed as Language Modeling problems, where the model performs autoregressive next-token prediction based on the input and historical context. 
To facilitate symbolic regression and align with the regression-oriented nature of these tasks, we adopt MSE as the training loss.

\noindent
\textbf{Symbolic Regression Details}
We apply PySR~\cite{cranmer2023interpretablemachinelearningscience}, a genetic-algorithm-based symbolic regression method, to identify the symbolic expression $\hat{f}$ that minimizes the prediction error on $\mathcal{D}$.
Specifically, we use PySR with its default hyperparameter settings; moreover, for nonlinear tasks, we introduce additional operators such as ReLU.
And to ensure reproducibility, we set the random seed to 42.
The overall runtime for extracting an algorithm from a model ranges from a few minutes to several tens of minutes.

% \paragraph{Non-State-Tracking Tasks (Token-Tagging).}
% These tasks require an output at position $t$ that is a strictly local function of a fixed-size window of inputs, with no dependency on an evolving historical state. We formulate these as Token-Tagging problems. Given an input sequence $X = (x_1, \dots, x_L)$, the model predicts an output sequence $Y = (y_1, \dots, y_L)$ in parallel. The loss is computed independently at each position:
% \begin{equation}
%     \mathcal{L}_{tag}(\theta) = \sum_{t=1}^{L} \text{Loss}(y_t, f(x_{t-k}, \dots, x_t))
% \end{equation}
% This design incentivizes the model to learn a single, position-invariant function applied uniformly across the sequence.

% \paragraph{State-Tracking Tasks (Language Modeling).}
% These tasks necessitate maintaining a state that evolves over the sequence, where the output at position $t$ depends on the state at $t-1$ and the current input $x_t$ (e.g., cumulative addition). We formulate these as Language Modeling (next-token prediction) problems. The core computation is a recursive state transition:
% \begin{equation}
%     s_t = f(s_{t-1}, x_t)
% \end{equation}
% By using next-token prediction, we externalize the state $s_t$ into the generated sequence. This simplifies the model's objective to learning the state transition function $f$, providing a clear and coherent target for symbolic extraction of the core algorithmic rule. The loss is calculated over shifted sequences:
% \begin{equation}
%     \mathcal{L}_{lm}(\theta) = \sum_{t=1}^{L} \text{Loss}
%     (y_t,  f(\mathbf{x}, y_{<t}))
% \end{equation}

\section{Robustness of the Discrete Transformer}
\label{app:robustness}

To evaluate the robustness of our framework under architectural variations, we investigate the impact of hyperparameters—specifically the number of layers, attention heads per layer, and sub-MLPs per layer—on both the convergence quality, measured by the MSE loss, and the complexity, quantified by the line count of the extracted program.
We conduct experiments on the \texttt{sum\_last2}, \texttt{parity\_last2}, and \texttt{spring} tasks, varying the number of layers in $\{1, 2, 3, 4\}$, attention heads per layer in $\{0, 1, 2, 4, 8\}$, and sub-MLPs per layer in $\{0, 1, 2, 4, 8\}$.

Our results exhibit a clear threshold effect in expressive capacity. 
Once the architecture satisfies the minimal functional requirements of the target task (e.g., \texttt{sum\_last2} requires at least one attention head to retrieve the token $x_{t-1}$), the Discrete Transformer reliably converges to a near-zero MSE loss (see Tables~\ref{tab:robustness_all_tasks_l12} and~\ref{tab:robustness_all_tasks_l34}).

Furthermore, we find that over-parameterization has two distinct effects in our framework. 
\begin{itemize}

\item First, excessive capacity increases the complexity of the extracted programs through structural redundancy: multiple modules may learn functionally equivalent behaviors, such as several attention heads retrieving the same offset token $x_{t-1}$. To examine whether this redundancy can be controlled, we introduce a simplicity bias by adding an $\ell_1$ regularization term with coefficient $0.001$ on the linear output head in the over-parameterized setting with 4 layers, 4 heads per layer, and 4 sub-MLPs per layer. On the \texttt{sum\_last2}, \texttt{parity\_last2}, and \texttt{spring} tasks, this reduces the extracted line counts from 78 to 22, 85 to 42, and 96 to 34, respectively. At the same time, the extraction RMSE remains low under regularization ($1.01\times 10^{-4}$, $8.71\times 10^{-3}$, and $1.14\times 10^{-4}$, respectively), suggesting that program complexity can be substantially reduced while maintaining high extraction fidelity.

\item Second, the presence of redundant modules expands the search space and introduces noise during the exploration phase. Although this can complicate optimization and reduce training stability, we find that a slower temperature annealing schedule mitigates the issue by giving redundant components more time to resolve competition and converge to a valid discrete solution. For example, on the \texttt{spring} task, using the same over-parameterized model with 4 layers, 4 heads per layer, and 4 sub-MLPs per layer, extending the annealing horizon from 25 to 50 and 100 epochs progressively reduces the training MSE loss from $3.69\times 10^{-3}$ to $2.35\times 10^{-6}$ and further to $3.94\times 10^{-7}$.

\end{itemize}

\begin{table*}[t]
\centering
\footnotesize
\setlength{\tabcolsep}{3pt}
\renewcommand{\arraystretch}{0.94}
\caption{Robustness tests across three tasks for architectures with one or two layers. For each task, SR denotes the success rate across three random seeds, Lines denotes the total lines of the extracted program, and RMSE denotes the mean RMSE across the three seeds. A dash indicates no valid program.}
\label{tab:robustness_all_tasks_l12}
\resizebox{\textwidth}{!}{%
\begin{tabular}{@{}ccc ccc ccc ccc@{}}
\toprule
\multirow{2}{*}{Layers} & \multirow{2}{*}{Heads} & \multirow{2}{*}{sub-MLPs}
& \multicolumn{3}{c}{\texttt{sum\_last2}}
& \multicolumn{3}{c}{\texttt{spring}}
& \multicolumn{3}{c}{\texttt{parity\_last2}} \\
\cmidrule(lr){4-6}\cmidrule(lr){7-9}\cmidrule(lr){10-12}
& & & SR & Lines & RMSE & SR & Lines & RMSE & SR & Lines & RMSE \\
\midrule
1 & 0 & 1 & 0/3 & -- & $3.16\times 10^{-1}$ & 0/3 & -- & $1.25\times 10^{1}$ & 0/3 & -- & $5.00\times 10^{-1}$ \\
1 & 0 & 2 & 0/3 & -- & $3.15\times 10^{-1}$ & 0/3 & -- & $1.25\times 10^{1}$ & 0/3 & -- & $5.00\times 10^{-1}$ \\
1 & 1 & 0 & 3/3 & 22 & $0$ & 0/3 & -- & $9.20\times 10^{0}$ & 0/3 & -- & $4.85\times 10^{-1}$ \\
1 & 1 & 1 & 3/3 & 25 & $4.02\times 10^{-6}$ & 0/3 & -- & $9.18\times 10^{0}$ & 3/3 & 25 & $1.79\times 10^{-7}$ \\
1 & 1 & 2 & 3/3 & 22 & $9.67\times 10^{-7}$ & 0/3 & -- & $6.06\times 10^{0}$ & 3/3 & 28 & $2.31\times 10^{-3}$ \\
1 & 2 & 0 & 3/3 & 22 & $5.25\times 10^{-5}$ & 3/3 & 28 & $0$ & 0/3 & -- & $4.85\times 10^{-1}$ \\
1 & 2 & 1 & 3/3 & 22 & $3.12\times 10^{-3}$ & 3/3 & 28 & $2.11\times 10^{-9}$ & 3/3 & 31 & $2.26\times 10^{-3}$ \\
1 & 2 & 2 & 3/3 & 28 & $1.43\times 10^{-6}$ & 3/3 & 28 & $7.41\times 10^{-7}$ & 3/3 & 34 & $2.31\times 10^{-3}$ \\
1 & 2 & 4 & 3/3 & 34 & $8.85\times 10^{-5}$ & 3/3 & 28 & $1.75\times 10^{-8}$ & 3/3 & 39 & $3.77\times 10^{-3}$ \\
1 & 2 & 8 & 3/3 & 22 & $4.08\times 10^{-6}$ & 3/3 & 28 & $3.69\times 10^{-6}$ & 3/3 & 52 & $3.79\times 10^{-4}$ \\
1 & 4 & 2 & 3/3 & 22 & $1.84\times 10^{-9}$ & 3/3 & 28 & $4.23\times 10^{-6}$ & 3/3 & 39 & $1.53\times 10^{-3}$ \\
1 & 4 & 4 & 3/3 & 36 & $5.00\times 10^{-7}$ & 3/3 & 28 & $9.90\times 10^{-6}$ & 3/3 & 28 & $2.06\times 10^{-3}$ \\
1 & 4 & 8 & 3/3 & 31 & $3.34\times 10^{-6}$ & 3/3 & 42 & $3.71\times 10^{-5}$ & 3/3 & 54 & $3.36\times 10^{-3}$ \\
1 & 8 & 2 & 3/3 & 31 & $6.31\times 10^{-7}$ & 3/3 & 34 & $1.30\times 10^{-6}$ & 3/3 & 62 & $1.79\times 10^{-3}$ \\
1 & 8 & 4 & 3/3 & 52 & $8.21\times 10^{-7}$ & 3/3 & 55 & $5.90\times 10^{-3}$ & 3/3 & 55 & $2.11\times 10^{-3}$ \\
1 & 8 & 8 & 3/3 & 41 & $6.34\times 10^{-6}$ & 3/3 & 71 & $6.85\times 10^{-4}$ & 3/3 & 78 & $4.57\times 10^{-3}$ \\
\midrule
2 & 2 & 2 & 3/3 & 47 & $8.27\times 10^{-3}$ & 3/3 & 46 & $1.28\times 10^{-3}$ & 3/3 & 47 & $3.48\times 10^{-6}$ \\
2 & 2 & 4 & 3/3 & 56 & $2.58\times 10^{-3}$ & 3/3 & 53 & $5.52\times 10^{-4}$ & 3/3 & 59 & $1.84\times 10^{-3}$ \\
2 & 2 & 8 & 3/3 & 53 & $1.74\times 10^{-5}$ & 3/3 & 67 & $3.78\times 10^{-4}$ & 3/3 & 52 & $1.10\times 10^{-3}$ \\
2 & 4 & 2 & 3/3 & 58 & $3.20\times 10^{-4}$ & 3/3 & 65 & $2.34\times 10^{-3}$ & 3/3 & 63 & $3.53\times 10^{-3}$ \\
2 & 4 & 4 & 3/3 & 60 & $1.31\times 10^{-4}$ & 3/3 & 74 & $6.88\times 10^{-3}$ & 3/3 & 59 & $1.39\times 10^{-3}$ \\
2 & 4 & 8 & 3/3 & 67 & $1.38\times 10^{-4}$ & 3/3 & 75 & $2.03\times 10^{-2}$ & 3/3 & 62 & $1.72\times 10^{-3}$ \\
2 & 8 & 2 & 3/3 & 73 & $9.81\times 10^{-5}$ & 3/3 & 63 & $2.14\times 10^{-3}$ & 3/3 & 80 & $9.58\times 10^{-4}$ \\
2 & 8 & 4 & 3/3 & 85 & $1.38\times 10^{-4}$ & 3/3 & 75 & $8.02\times 10^{-4}$ & 3/3 & 85 & $4.57\times 10^{-3}$ \\
2 & 8 & 8 & 3/3 & 72 & $1.09\times 10^{-4}$ & 3/3 & 86 & $3.89\times 10^{-3}$ & 3/3 & 68 & $3.44\times 10^{-3}$ \\
\bottomrule
\end{tabular}%
}
\end{table*}

\begin{table*}[t]
\centering
\footnotesize
\setlength{\tabcolsep}{3pt}
\renewcommand{\arraystretch}{0.94}
\caption{Robustness tests across three tasks for architectures with three or four layers. The notation follows Table~\ref{tab:robustness_all_tasks_l12}.}
\label{tab:robustness_all_tasks_l34}
\resizebox{\textwidth}{!}{%
\begin{tabular}{@{}ccc ccc ccc ccc@{}}
\toprule
\multirow{2}{*}{Layers} & \multirow{2}{*}{Heads} & \multirow{2}{*}{sub-MLPs}
& \multicolumn{3}{c}{\texttt{sum\_last2}}
& \multicolumn{3}{c}{\texttt{spring}}
& \multicolumn{3}{c}{\texttt{parity\_last2}} \\
\cmidrule(lr){4-6}\cmidrule(lr){7-9}\cmidrule(lr){10-12}
& & & SR & Lines & RMSE & SR & Lines & RMSE & SR & Lines & RMSE \\
\midrule
3 & 2 & 2 & 3/3 & 60 & $4.46\times 10^{-5}$ & 3/3 & 53 & $4.05\times 10^{-4}$ & 3/3 & 52 & $1.14\times 10^{-4}$ \\
3 & 2 & 4 & 3/3 & 55 & $1.54\times 10^{-5}$ & 3/3 & 65 & $6.53\times 10^{-3}$ & 3/3 & 66 & $4.41\times 10^{-3}$ \\
3 & 2 & 8 & 3/3 & 82 & $9.45\times 10^{-4}$ & 3/3 & 82 & $4.86\times 10^{-3}$ & 3/3 & 74 & $5.55\times 10^{-4}$ \\
3 & 4 & 2 & 3/3 & 71 & $3.71\times 10^{-4}$ & 3/3 & 75 & $4.64\times 10^{-3}$ & 3/3 & 58 & $1.48\times 10^{-4}$ \\
3 & 4 & 4 & 3/3 & 61 & $5.70\times 10^{-5}$ & 3/3 & 82 & $2.12\times 10^{-3}$ & 3/3 & 81 & $8.05\times 10^{-5}$ \\
3 & 4 & 8 & 3/3 & 82 & $1.92\times 10^{-4}$ & 3/3 & 76 & $1.52\times 10^{-3}$ & 3/3 & 79 & $4.62\times 10^{-3}$ \\
3 & 8 & 2 & 3/3 & 86 & $1.96\times 10^{-3}$ & 3/3 & 92 & $7.65\times 10^{-4}$ & 3/3 & 77 & $3.96\times 10^{-3}$ \\
3 & 8 & 4 & 3/3 & 85 & $4.90\times 10^{-4}$ & 3/3 & 99 & $4.39\times 10^{-3}$ & 3/3 & 70 & $3.48\times 10^{-3}$ \\
3 & 8 & 8 & 3/3 & 74 & $4.44\times 10^{-4}$ & 3/3 & 101 & $1.05\times 10^{-3}$ & 3/3 & 76 & $2.31\times 10^{-3}$ \\
\midrule
4 & 2 & 2 & 3/3 & 64 & $1.84\times 10^{-1}$ & 3/3 & 64 & $1.20\times 10^{-3}$ & 3/3 & 68 & $2.45\times 10^{-3}$ \\
4 & 2 & 4 & 3/3 & 83 & $3.06\times 10^{-4}$ & 3/3 & 86 & $1.27\times 10^{-3}$ & 3/3 & 82 & $1.47\times 10^{-4}$ \\
4 & 2 & 8 & 3/3 & 108 & $1.40\times 10^{-3}$ & 3/3 & 88 & $4.22\times 10^{-3}$ & 3/3 & 62 & $4.50\times 10^{-4}$ \\
4 & 4 & 2 & 3/3 & 77 & $1.32\times 10^{-4}$ & 3/3 & 64 & $5.28\times 10^{-4}$ & 3/3 & 70 & $3.14\times 10^{-3}$ \\
4 & 4 & 4 & 3/3 & 86 & $1.11\times 10^{-3}$ & 3/3 & 96 & $3.04\times 10^{-3}$ & 3/3 & 85 & $3.26\times 10^{-3}$ \\
4 & 4 & 8 & 3/3 & 89 & $4.89\times 10^{-4}$ & 3/3 & 124 & $9.06\times 10^{-3}$ & 3/3 & 75 & $4.85\times 10^{-3}$ \\
4 & 8 & 2 & 3/3 & 78 & $1.33\times 10^{-1}$ & 3/3 & 98 & $4.83\times 10^{-4}$ & 3/3 & 79 & $4.45\times 10^{-3}$ \\
4 & 8 & 4 & 3/3 & 103 & $5.49\times 10^{-4}$ & 3/3 & 116 & $2.40\times 10^{-3}$ & 3/3 & 70 & $3.17\times 10^{-3}$ \\
4 & 8 & 8 & 3/3 & 123 & $5.83\times 10^{-4}$ & 3/3 & 115 & $2.06\times 10^{-3}$ & 3/3 & 102 & $2.32\times 10^{-3}$ \\
\bottomrule
\end{tabular}%
}
\end{table*}

% Our results reveal a distinct threshold behavior regarding expressivity.
% Provided the architecture meets the minimal functional requirements of the target task—for instance, \texttt{sum\_last2} theoretically necessitates at least one attention head to retrieve the token $x_{t-1}$—the Discrete Transformer consistently converges to a near-zero MSE loss.
% However, over-parameterization introduces significant dynamics in the discrete search process.
% We observe that excessive capacity tends to increase the length of the extracted programs.
% This inflation arises from structural redundancy, where the model allocates multiple modules to perform identical functions (e.g., several attention heads learning the same retrieval pattern for $x_{t-1}$).
% Furthermore, the introduction of redundant modules expands the search space and introduces noise during the exploration phase.
% While this poses a challenge to the optimization stability, we find that it can be effectively addressed by adopting a slower temperature annealing schedule, allowing the model sufficient time to resolve the competition between redundant components and settle into a valid discrete solution.

\section{Additional Synthesized Programs}
\label{sec: synthesized programs}

In this section, we provide additional synthesized programs.

\begin{figure*}[t]
\centering
\begin{minipage}[t]{0.48\textwidth}
    \begin{lstlisting}
import numpy as np
def maximum_prev2(input_seq):
    # V0_input
    input_arr = np.array(input_seq, dtype=float)
    seq_len = input_arr.shape[0]
    V0 = input_arr[:, 0]
    # V1_Attn_L0H0
    V1 = np.zeros(seq_len)
    # windowed max
    for t in range(seq_len):
        V1[t] = np.max(V0[max(0, t-1): t+1])
    # V2_Attn_L0H1
    V2 = np.zeros(seq_len)
    # windowed max
    for t in range(seq_len):
        V2[t] = np.max(V0[max(0, t-1): t+1])
    # Output Head
    output = 1.07 * V1 + -0.07 * V2
    return output
    \end{lstlisting}
% 可以在这里加单独的 caption
% \captionof{figure}{Small caption if needed}
% \end{minipage}

    % \vspace{0.2cm} % 上下两个代码块之间的间距

% \begin{minipage}[t]{\columnwidth}
    \begin{lstlisting}
import numpy as np
def minimum_prev2(input_seq):
    # V0_input
    input_arr = np.array(input_seq, dtype=float)
    seq_len = input_arr.shape[0]
    V0 = input_arr[:, 0]
    # V1_Attn_L0H0
    V1 = np.zeros(seq_len)
    # windowed min
    for t in range(seq_len):
        V1[t] = np.min(V0[max(0, t-1): t+1])
    # Output Head
    output = 1.00 * V1
    return output
    \end{lstlisting}
\end{minipage}
\hfill % 左右两列之间的弹簧空白
% --- 右侧列 (包含右上) ---
\begin{minipage}[t]{0.48\textwidth}
    \begin{lstlisting}
import numpy as np
def spring(input_seq):
    # V0_input
    input_arr = np.array(input_seq, dtype=float)
    seq_len = input_arr.shape[0]
    V0 = input_arr[:, 0]
    # V1_Attn_L0H0
    V1 = np.zeros(seq_len)
    # Fixed offset 9
    V1[9:] = V0[:-9]
    # V2_Attn_L0H1
    V2 = np.zeros(seq_len)
    # Fixed offset 10
    V2[10:] = V0[:-10]
    # V3_MLP_L0M0
    V3 = np.full(seq_len, 0.02)
    # V4_MLP_L0M1
    V4 = np.full(seq_len, -0.11)
    # V5_Attn_L1H0
    V5 = np.zeros(seq_len)
    # Fixed offset 2
    V5[2:] = V0[:-2]
    # V6_Attn_L1H1
    V6 = np.zeros(seq_len)
    # Fixed offset 2
    V6[2:] = V4[:-2]
    # V7_MLP_L1M0
    V7 = np.full(seq_len, -0.03)
    # Output Head
    output = 1.00 * V1 + 1.00 * V2 + -1.00 * V5 + 0.03 * V7 + -0.01 * V6
    return output
    \end{lstlisting}
\end{minipage}
% \vspace{-2mm} % 调整 caption 与代码块的垂直距离
% \caption{Intervened synthesized programs for \texttt{maximum\_prev2} (Left Top), \texttt{minimum\_prev2} (Left Bottom), and \texttt{spring} (Right).
% }
\caption{
Intervened synthesized programs revealing alternative logical pathways.
Left: When MLP-based arithmetic is prohibited, the model solves \texttt{maximum\_prev2} and \texttt{minimum\_prev2} by shifting to a pure attention mechanism. The extracted code shows explicit \textit{Windowed Extrema} attention patterns.
Right: For the \texttt{spring} task, intervention in the information flow (masking recent history) forces the model to learn a high-order recurrence relation, validating the model's ability to uncover multiple equivalent algorithms for the same data distribution.
}

\label{fig:code_comparison_after_constraint}
\end{figure*}

\begin{figure*}[t]
\centering
\begin{minipage}[t]{0.48\textwidth}
    \begin{lstlisting}
import numpy as np
def sum_last(input_seq):
    # V0_input
    input_arr = np.array(input_seq, dtype=float)
    seq_len = input_arr.shape[0]
    V0 = input_arr[:, 0]
    # V1_Attn_L0H0
    V1 = np.zeros(seq_len)
    # Fixed offset 9
    V1[9:] = V0[:-9]
    # Output Head
    output = 1.00 * V0 + 1.00 * V1
    return output
    \end{lstlisting}
% 可以在这里加单独的 caption
% \captionof{figure}{Small caption if needed}
% \end{minipage}

    % \vspace{0.2cm} % 上下两个代码块之间的间距

% \begin{minipage}[t]{\columnwidth}
    \begin{lstlisting}
import numpy as np
def bitwise_and(input_seq):
    # V0_input, V1_input
    input_arr = np.array(input_seq, dtype=float)
    seq_len = input_arr.shape[0]
    V0 = input_arr[:, 0]
    V1 = input_arr[:, 1]
    # V4_MLP_L0M0
    V4 = np.full(seq_len, 0.02)
    # V5_MLP_L0M1
    V5 = ((((V0 - 0.49) * 3.28) / (V1 + -0.52)) - 3.06) * 0.19
    # Output Head
    output = 0.49 * V1 + 0.48 * V0 + 0.40 * V5 + 0.02 * V4
    return output
    \end{lstlisting}
\end{minipage}
\hfill % 左右两列之间的弹簧空白
% --- 右侧列 (包含右上) ---
\begin{minipage}[t]{0.48\textwidth}
    \begin{lstlisting}
import numpy as np
def add_mod_3(input_seq):
    # V0_input
    input_arr = np.array(input_seq, dtype=float)
    seq_len = input_arr.shape[0]
    V0 = input_arr[:, 0]
    # V1_Attn_L0H0
    V1 = np.zeros(seq_len)
    # Fixed offset 9
    V1[9:] = V0[:-9]
    # V2_Attn_L0H1
    V2 = np.zeros(seq_len)
    # Fixed offset 9
    V2[9:] = V0[:-9]
    # V3_MLP_L0M0
    V3 = ((-0.70 / (((V0 + V1) + 0.59) * -0.56)) - 1.69) + (1.05 / (V0 + (V1 - 2.53)))
    # Output Head
    output = -0.53 * V3 + 0.47 * V2 + -0.32 * V1 + 0.16 * V0
    return output
    \end{lstlisting}
\end{minipage}
% \vspace{-2mm} % 调整 caption 与代码块的垂直距离
% \caption{Intervened synthesized programs for \texttt{maximum\_prev2} (Left Top), \texttt{minimum\_prev2} (Left Bottom), and \texttt{spring} (Right).
% }
\caption{
Synthesized programs for \texttt{sum\_last} (Top Left), \texttt{bitwise\_and} (Bottom Left), and \texttt{add\_mod\_3} (Right).
}
\label{fig:non-linear}
\end{figure*}

\begin{figure*}[t]
\centering
% --- 左侧列 (包含左上和左下) ---
\begin{minipage}[t]{0.48\textwidth}
    % 1. 左上: newton_freebody
    \begin{lstlisting}
import numpy as np
def freebody(input_seq):
    # V0_input
    input_arr = np.array(input_seq, dtype=float)
    seq_len = input_arr.shape[0]
    V0 = input_arr[:, 0]
    # V1_Attn_L0H0
    V1 = np.zeros(seq_len)
    # Fixed offset 9
    V1[9:] = V0[:-9]
    # V2_Attn_L0H1
    V2 = np.zeros(seq_len)
    # Fixed offset 10
    V2[10:] = V0[:-10]
    # V4_MLP_L0M1
    V4 = -0.26*V0 + 0.04
    # V5_MLP_L0M2
    V5 = -0.21*V0 + -0.04
    # V6_Attn_L1H0
    V6 = np.zeros(seq_len)
    # Fixed offset 2
    V6[2:] = V4[:-2]
    # V7_Attn_L1H1
    V7 = np.zeros(seq_len)
    # Fixed offset 2
    V7[2:] = V5[:-2]
    # Output Head
    output = 1.19 * V0 + 1.06 * V7 + 1.06 * V6 + 1.00 * V1 + -0.65 * V4 + -0.65 * V5 + 0.50 * V2
    return output
    \end{lstlisting}

    % \vspace{0.2cm} % 上下两个代码块之间的间距

    % 2. 左下: newton_spring
    \begin{lstlisting}
import numpy as np
def spring(input_seq):
    # V0_input
    input_arr = np.array(input_seq, dtype=float)
    seq_len = input_arr.shape[0]
    V0 = input_arr[:, 0]
    # V1_Attn_L0H0
    V1 = np.zeros(seq_len)
    # Fixed offset 9
    V1[9:] = V0[:-9]
    # V2_Attn_L0H1
    V2 = np.zeros(seq_len)
    # Fixed offset 1
    V2[1:] = V0[:-1]
    # V4_MLP_L0M1
    V4 = (V2 + ((V1 - ((V1 + 2.26) + (V2 - 1.54))) * 2.00)) + (V2 + 1.43)
    # Output Head
    output = -1.00 * V2 + 1.00 * V0 + 1.00 * V1 + 0.04 * V4
    return output
    \end{lstlisting}
\end{minipage}
\hfill % 左右两列之间的弹簧空白
% --- 右侧列 (包含右上) ---
\begin{minipage}[t]{0.48\textwidth}
    % 3. 右上: newton_gravity
    \begin{lstlisting}
import numpy as np
def gravity(input_seq):
    # V0_input
    input_arr = np.array(input_seq, dtype=float)
    seq_len = input_arr.shape[0]
    V0 = input_arr[:, 0]
    # V1_Attn_L0H0
    V1 = np.zeros(seq_len)
    # Fixed offset 9
    V1[9:] = V0[:-9]
    # V2_Attn_L0H1
    V2 = np.zeros(seq_len)
    # Fixed offset 10
    V2[10:] = V0[:-10]
    # V4_MLP_L0M1
    V4 = 0.00
    # V5_MLP_L0M2
    V5 = -0.40*V0 + -3.05
    # V6_Attn_L1H0
    V6 = np.zeros(seq_len)
    # Fixed offset 2
    V6[2:] = V5[:-2]
    # V7_Attn_L1H1
    V7 = np.zeros(seq_len)
    # Fixed offset 2
    V7[2:] = V4[:-2]
    # Output Head
    output = 1.24 * V6 + 1.20 * V0 + 1.00 * V1 + -0.74 * V5 + 0.50 * V2 + -0.01 * V4 + 0.01 * V7
    return output
    \end{lstlisting}

\end{minipage}

% \vspace{-2mm}
\caption{
Synthesized programs for \texttt{freebody} (Top Left), \texttt{spring} (Bottom Left), and \texttt{gravity} (Right).
% By coordinating the Numerical Attention, Numerical MLP and linear output head, the model effectively identifies essential computational variables and constructs the correct computation graph.
% This confirms the model's capability to learn underlying physical laws and demonstrates its potential for modeling complex physical processes.
}
\label{fig:newton}
\end{figure*}

\begin{table*}[t]
    \centering
    \caption{Performance and parameters of the Discrete Transformer on the algorithmic benchmark. The columns Layers, Heads, and sub-MLPs correspond to the number of layers, the number of attention heads per layer, and the number of sub-MLPs per layer, respectively. Loss values are averaged MSE over three random seeds.}
    \label{tab:all_experiment_results}
    \begin{small}
    \begin{tabular}{llcccc}
        \toprule
        \textbf{Task Name} & \textbf{Description} & \textbf{Layers} & \textbf{Heads} & \textbf{sub-MLPs} & \textbf{Train Loss} \\
        \midrule
        \multicolumn{6}{l}{\textit{Linear Arithmetic}} \\
        \texttt{sum\_last2} & Sum of the last two numbers & 1 & 2 & 2 & $8.82 \times 10^{-9\phantom{0}}$\\
        \texttt{diff\_last2} & Difference of the last two numbers & 1 & 2 & 2 & $6.59 \times 10^{-16}$\\
        \texttt{sum} & Cumulative sum of the sequence & 1 & 2 & 2 & $1.71 \times 10^{-14}$\\
        \texttt{div\_3} & Binary division by 3 & 1 & 2 & 2 & $2.20 \times 10^{-1\phantom{0}}$\\
        \texttt{freebody} & Simulate freebody dynamics & 2 & 2 & 3 & $9.57 \times 10^{-11}$\\
        \texttt{gravity} & Simulate gravity dynamics & 2 & 2 & 3 & $2.34 \times 10^{-11}$\\
        \texttt{spring} & Simulate spring dynamics & 1 & 2 & 2 & $1.41 \times 10^{-7\phantom{0}}$\\
        \texttt{magnetic} & Simulate magnetic dynamics & 2 & 2 & 3 & $1.12 \times 10^{2\phantom{0}\phantom{-}}$\\
        \midrule
        \multicolumn{6}{l}{\textit{Non-Linear Composition}} \\
        \texttt{parity\_last2} & Parity of the last two numbers & 1 & 2 & 2 & $4.87 \times 10^{-13}$\\
        \texttt{maximum\_prev2} & Maximum of the last two numbers & 1 & 2 & 2 & $1.11 \times 10^{-6\phantom{0}}$\\
        \texttt{minimum\_prev2} & Minimum of the last two numbers & 1 & 2 & 2 & $1.64 \times 10^{-5\phantom{0}}$\\
        \texttt{unique\_prev2} & 1 if last two numbers are equal & 1 & 2 & 2 & $4.19 \times 10^{-14}$\\
        \texttt{bitwise\_and} & Bitwise AND & 1 & 2 & 2 & $1.10 \times 10^{-18}$\\
        \texttt{bitwise\_or} & Bitwise OR & 1 & 2 & 2 & $5.66 \times 10^{-4\phantom{0}}$\\
        \texttt{bitwise\_not} & Bitwise NOT & 1 & 2 & 2 & $1.44 \times 10^{-17}$\\
        \texttt{bitwise\_xor} & Bitwise XOR & 1 & 2 & 2 & $6.74 \times 10^{-20}$\\
        \texttt{abs} & Abs value of the number & 1 & 2 & 2 & $2.69 \times 10^{-14}$\\
        \texttt{abs\_of\_diff} & Abs difference of last two numbers & 1 & 2 & 2 & $9.72 \times 10^{-9\phantom{0}}$\\
        \texttt{parity} & Cumulative parity & 1 & 2 & 2 & $8.15 \times 10^{-17}$\\
        \texttt{parity\_zeros} & Cumulative parity of number of zeros & 1 & 2 & 2 & $2.25 \times 10^{-8\phantom{0}}$\\
        \texttt{add\_mod\_3} & Cumulative sum modulo 3 & 1 & 2 & 2 & $9.72 \times 10^{-13}$\\        
        \texttt{bit\_dot\_prod\_mod\_2} & Cumulative dot product of bits mod 2 & 1 & 2 & 2 & $5.55 \times 10^{-4\phantom{0}}$\\
        \texttt{bit\_addition} & Binary addition & 2 & 4 & 4 & $2.49 \times 10^{-2\phantom{0}}$\\
        \texttt{balanced\_parenthesis} & 1 if parentheses are balanced & 1 & 2 & 2 & $2.08 \times 10^{-2\phantom{0}}$\\
        \midrule
        \multicolumn{6}{l}{\textit{Continuous Dynamics}} \\
        \texttt{exponential\_moving\_average} & Exponential moving average & 1 & 2 & 2 & $1.62 \times 10^{-13}$ \\
        \texttt{free\_fall\_height} & Free-fall height & 1 & 1 & 1 & $8.53 \times 10^{-5\phantom{0}}$ \\
        \texttt{linear\_drop} & Linear drop & 2 & 2 & 2 & $2.52 \times 10^{-7\phantom{0}}$ \\
        \texttt{quadratic\_drop} & Quadratic drop & 2 & 2 & 2 & $8.74 \times 10^{-7\phantom{0}}$ \\
        \texttt{damped\_harmonic\_oscillator} & Damped harmonic oscillator & 1 & 2 & 2 & $8.84 \times 10^{-8\phantom{0}}$ \\
        \bottomrule
    \end{tabular}
    \end{small}

\end{table*}


% You can have as much text here as you want. The main body must be at most $8$
% pages long. For the final version, one more page can be added. If you want, you
% can use an appendix like this one.

% The $\mathtt{\backslash onecolumn}$ command above can be kept in place if you
% prefer a one-column appendix, or can be removed if you prefer a two-column
% appendix.  Apart from this possible change, the style (font size, spacing,
% margins, page numbering, etc.) should be kept the same as the main body.
% %%%%%%%%%%%%%%%%%%%%%%%%%%%%%%%%%%%%%%%%%%%%%%%%%%%%%%%%%%%%%%%%%%%%%%%%%%%%%%%
% %%%%%%%%%%%%%%%%%%%%%%%%%%%%%%%%%%%%%%%%%%%%%%%%%%%%%%%%%%%%%%%%%%%%%%%%%%%%%%%

\end{document}